\documentclass[journal]{IEEEtran}

\usepackage{amsfonts}
\usepackage{microtype}
\usepackage{nomencl}
\usepackage{graphicx}
\usepackage{subfigure}
\usepackage{booktabs}
\usepackage{pdfpages}

\usepackage{multirow}
\usepackage{mathtools,cases}
\usepackage{xcolor}
\usepackage{hyperref}
\usepackage{etoolbox} % create groups for nomenclature

\def\nmo{N$-$1}%

\makenomenclature

\renewcommand\nomgroup[1]{\vspace{3mm}
  \item[\bfseries
  \ifstrequal{#1}{S}{Sets}{
  \ifstrequal{#1}{V}{\hspace*{-2mm}Variables}{
  \ifstrequal{#1}{F}{\hspace*{-4mm} Functions and operators}{}}}]
\hspace*{-\leftmargin}\vspace{3mm}
}

\newcommand{\Graph}{\mathbb{G}}
\newcommand{\Node}{\mathcal{N}}
\newcommand{\Edge}{\mathcal{E}}
\newcommand{\Load}{\mathcal{L}}
\newcommand{\Generator}{\mathcal{G}}

\begin{document}

% set url style
\urlstyle{tt}

\title{Leveraging power grid topology in machine learning assisted optimal power flow}

\author{Thomas~Falconer
        and~Letif~Mones% <-this % stops a space
\thanks{All authors are with Invenia Labs, 95 Regent Street, Cambridge, CB2 1AW, United Kingdom (e-mails: \{firstname.lastname\}@invenialabs.co.uk).}% <-this % stops a space
}

\maketitle

\begin{abstract}
Machine learning assisted optimal power flow (OPF) aims to reduce the computational complexity of these non-linear and non-convex constrained optimization problems by consigning expensive (online) optimization to offline training. The majority of work in this area typically employs fully connected neural networks (FCNN). However, recently convolutional (CNN) and graph (GNN) neural networks have also been investigated, in effort to exploit topological information within the power grid. Although promising results have been obtained, there lacks a systematic comparison between these architectures throughout literature. Accordingly, we introduce a concise framework for generalizing methods for machine learning assisted OPF and assess the performance of a variety of FCNN, CNN and GNN models for two fundamental approaches in this domain: regression (predicting optimal generator set-points) and classification (predicting the active set of constraints). For several synthetic power grids with interconnected utilities, we show that locality properties between feature and target variables are scarce and subsequently demonstrate marginal utility of applying CNN and GNN architectures compared to FCNN for a fixed grid topology. However, with variable topology (for instance, modeling transmission line contingency), GNN models are able to straightforwardly take the change of topological information into account and outperform both FCNN and CNN models.
\end{abstract}


\IEEEpeerreviewmaketitle

%\mbox{} % white space

% sets
\nomenclature[S]{$\mathcal{N}$, $\mathcal{E}$}{Sets of nodes (vertices) and edges that define an undirected graph, $\Graph$, respectively}
\nomenclature[S]{$\mathcal{M}$}{Full set of neural network models for which predictive performance is assessed.}
\nomenclature[S]{$\mathcal{C}^{\mathrm{E}}$, $\mathcal{C}^{\mathrm{I}}$}{Full sets of equality and inequality constraints for a particular OPF problem, respectively.}
\nomenclature[S]{$\mathcal{A}$}{Set of active inequality constraints (those satisfied with equality at the optimal point).}
\nomenclature[S]{$\mathcal{F}_{\Phi}$}{Set of feasible points for the optimization variables.}
\nomenclature[S]{$\Omega$}{Abstract set representing the OPF operator domain.}
\nomenclature[S]{$\mathcal{V}$}{Set of violated inequality constraints associated with a vector of optimization variables, $y$.}
\nomenclature[S]{$\sigma$}{Set of hyperparameters used to define neural network architectures.}
\nomenclature[S]{$\theta$}{Set of neural network parameters optimized during the model training process.}

% variables
\nomenclature[V]{$x$}{Vector of grid parameters (e.g. active and reactive power components of loads).}
\nomenclature[V]{$y$}{Vector of primal variables (e.g. voltage magnitudes and active power component of generator injections).}
\nomenclature[V]{$z$}{Vector of dual variables (Lagrangian multipliers) of the associated equality and inequality constraints.}
\nomenclature[V]{$P_g$, $P_l$}{Power injection and withdrawal for a particular generator and load, respectively (active power components).}
\nomenclature[V]{$V_m$}{Bus voltage magnitude.}
\nomenclature[V]{$Z_{ij}$}{Impedance of transmission line between bus $i$ and bus $j$.}

% functions
\nomenclature[F]{$\Phi$, $\Psi$}{OPF operators that map grid parameters to optimal values of the primal variables and both primal and dual variables, respectively.}
\nomenclature[F]{$F$}{OPF function introduced to simplify notation of the related operator whereby only grid parameters vary.}
\nomenclature[F]{$l$}{Loss function used to optimize neural network parameters, $\theta$.}
\nomenclature[F]{$f$}{Objective function of a particular OPF problem.}
\nomenclature{$\mathbb{G}$}{An undirected graph with $|\mathcal{N}|$ number of nodes and $\mathcal{E}$ number of edges.}

\printnomenclature

\section{Introduction}
\label{sec:intro}

\IEEEPARstart{O}{ptimal} power flow (OPF) is an umbrella term for a family of constrained optimization problems that govern electricity market dynamics and facilitate effective planning and operation of modern power systems \cite[p.~514]{Wood2014}. Classical OPF (AC-OPF) formulates a non-linear and non-convex economic dispatch model, which minimizes the cost of generator scheduling subject to either (or both) operation and security constraints of the grid \cite{Billinton1994}. By virtue of competitive efficiency, optimal schedules are typically found using interior-point methods \cite{Wachter2006}. However, the required computation of the Hessian (second-order derivatives) of the Lagrangian at each optimization step renders a super-linear time complexity, thus large-scale systems can be prohibitively slow to solve.

This computational constraint gives rise to several challenges for independent system operators (ISOs): (1) variable inclusion of certain generators (i.e. unit commitment) invokes binary variables in the optimization model, thereby forming a mixed-integer, non-linear program (known to be NP-hard), exacerbating computational costs \cite{Castillo2016}; (2) the standard requirement for operators to satisfy \nmo{} security constraints (i.e. account for all contingency events where a single grid component fails) renders a much larger-scale problem, increasing the time complexity even further \cite{Rahman2020}; and lastly (3) modeling uncertainty in the supply-demand equilibrium induced by stochastic renewable generation requires methods such as scenario based Monte-Carlo simulation \cite{Mezghani2020}, which necessitates sequential OPF solutions at rates unattainable by conventional algorithms. 

To overcome these challenges, ISOs often resort to simplified OPF models by utilizing convex relaxations \cite{Low2014} or linearizations \cite{Bolognani2015, Bernstein2017} such as the widely adopted DC-OPF model \cite{Cain2012}. With considerably less control variables and constraints, DC-OPF can be solved very efficiently using interior-point or simplex methods \cite[p.~224]{Vonmeier2006}.
However, as DC-OPF solutions are in fact never feasible with respect to the full problem \cite{Baker2020}, set-points need to be found iteratively by manually updating the solution until convergence \cite[p.~14]{FERC2011} -- hence DC-OPF is predisposed to sub-optimal generator scheduling.

\begin{figure}[!ht]
    \centerline{\includegraphics[width=\columnwidth]{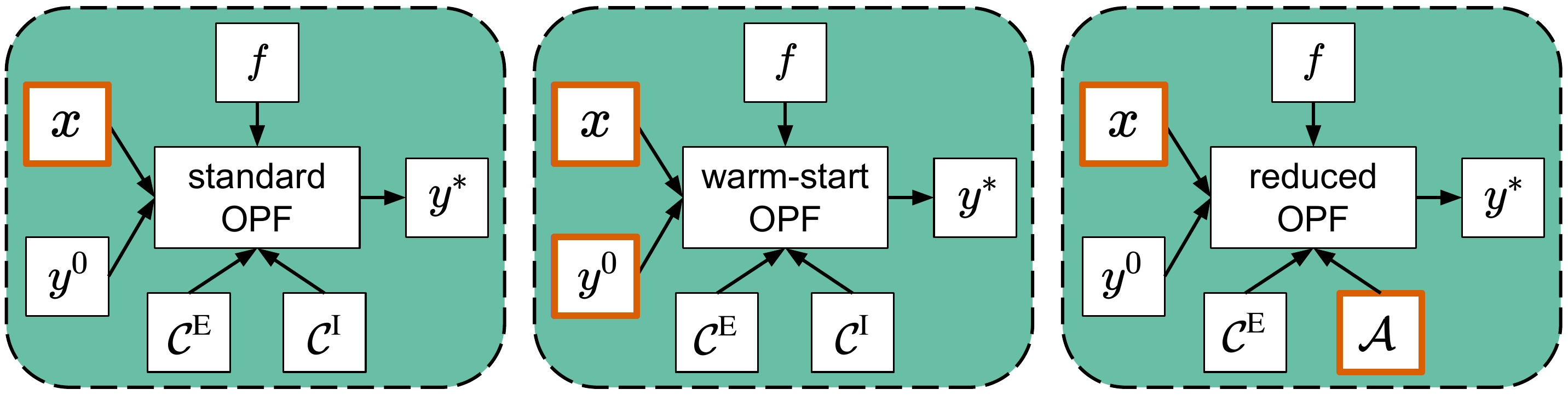}}
    \caption{Strategies for solving OPF with interior-point methods: standard (left), warm-start (center) and reduced (right) problems. $x$ and $y$ are the vectors of grid parameters and optimization variables, respectively, $f$ is the objective function, $\mathcal{C}^{\mathrm{E}}$ and $\mathcal{C}^{\mathrm{I}}$ denote the sets of equality and inequality constraints, and $\mathcal{A} \subseteq \mathcal{C}^{\mathrm{I}}$ is the active subset of the inequality constraints. Typical varying arguments are highlighted in orange, whilst remaining arguments are potentially fixed.}
    \label{fig:ml_opf_strategies}
\end{figure}

In practice, ISOs typically leverage additional information about the grid in attempt to obtain solutions more efficiently. For instance, given the (reasonable) assumption that comparable grid states will correspond to neighbouring points in solution space, one can use the known solution to a similar problem as the starting value for the optimization variables of another problem -- a so-called \emph{warm-start} (Figure~\ref{fig:ml_opf_strategies}, center panel) --, rendering considerably faster convergence compared to arbitrary initialisation \cite{Shahzad2010}. Alternatively, ISOs can capitalize on the observation that only a fraction of inequality constraints are actually binding at the optimal point \cite{Zhou2011}, hence one can remove a large number of constraints from the mathematical model and formulate an equivalent, but significantly cheaper, \emph{reduced problem} \cite{Roald2019} (Figure~\ref{fig:ml_opf_strategies}, right panel). Security risks associated with the omission of violated constraints from the reduced problem can be mitigated by iteratively solving the reduced OPF and updating the active set until all constraints of the full problem are satisfied \cite{Xingwang2009}.

\subsection{Machine learning assisted OPF}
A compelling new area of research borne from the machine learning community attempts to alleviate reliance on subpar OPF frameworks by fitting an estimator functions on historical data. The estimators are typically neural networks (NNs) owed to their demonstrated ability to model complex non-linear relationships with negligible online computation \cite{LeCun2015}. This makes it possible to obtain predictions in real-time, thereby shifting the computational expense from online optimization to offline training -- and the trained model can remain sufficient for a period of time, requiring only occasional re-training.

Most of the NN-based methods for machine learning assisted OPF can be generalized as one of two approaches: (1) \textit{end-to-end} (or direct) models, where an estimator function is used to learn a direct mapping between the grid parameters and the optimal OPF solution; and (2) \textit{hybrid} (or indirect) techniques -- a two-step approach whereby an estimator function first maps the grid parameters to some quantities, which are subsequently used as inputs to an optimization problem to find a (possibly exact) solution. Based on the actual target type, these methods can be further categorized depending on the type of predicted quantity: regression or classification. 

\subsubsection{Regression}
By inferring OPF solutions directly, end-to-end regression methods bypass conventional solvers altogether, offering the greatest (online) computational gains \cite{Guha2019}. However, since OPF is a constrained optimization problem, the optimal solution is not necessarily a smooth function of the inputs: changes of the binding status of constraints can lead to abrupt changes of the optimal solution. Since the number of unique sets of binding constraints increases exponentially with system size, this approach requires training on relatively large data sets in order to obtain sufficient accuracy \cite{Fioretto2019}. Moreover, there is no guarantee that the inferred solution is feasible, and violation of important constraints poses severe security risks to the grid.

\begin{figure}[!ht]
    \centerline{\includegraphics[width=\columnwidth]{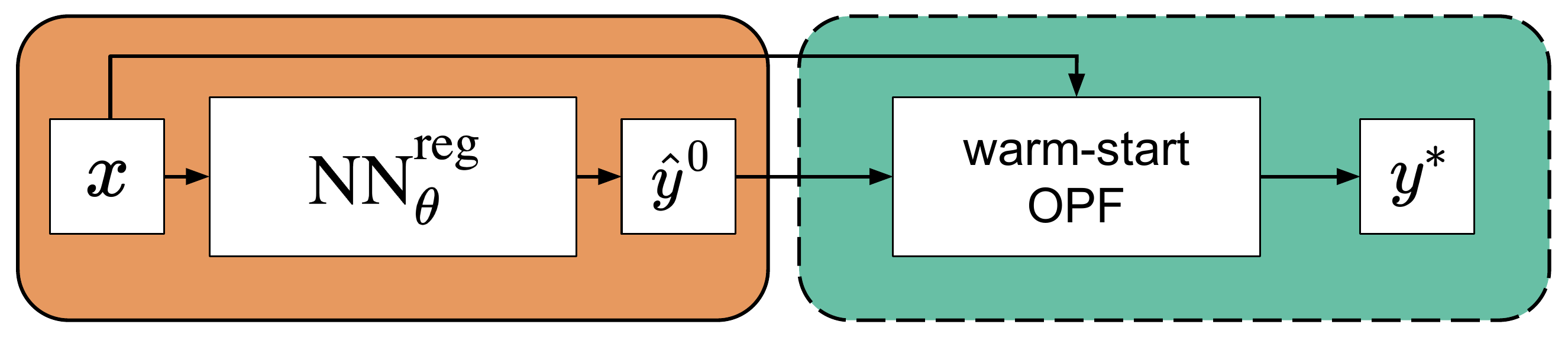}}
    \caption{Flowchart of the warm-start method (green panel) combined with a NN regressor (orange panel). For clarity, default arguments of the OPF operator are omitted.}
    \label{fig:warm_start}
\end{figure}

Instead, one can adopt a hybrid approach whereby the inferred solution of the end-to-end method is used to initialize an interior-point solver (i.e. a warm-start), which provides an optimal solution to an optimization problem equivalent to the original one (Figure~\ref{fig:warm_start}). Compared to default heuristics used in the conventional optimization method, an accurate initial point could theoretically reduce the number of required iterations (and so the computational cost) to reach the optimal point \cite{Baker2019}. However, as discussed in \cite{Robson2020}, there are several practical issues which could arise, leading to limited computational gain for this technique. 

\subsubsection{Classification}
An alternative hybrid approach leverages the aforementioned technique of formulating a reduced problem by removing non-binding inequality constraints from the mathematical model. A NN classifier is first used to predict the active set of constraints by either (1) identifying all distinct active sets in the training data and using a multi-class classifier to map the features accordingly \cite{Misra2018}; or (2) by predicting the binding status of each inequality constraint using a binary multi-label classifier \cite{Robson2020}. Since the number of active sets increases exponentially with system size \cite{Deka2019}, the latter approach may be computationally favourable for larger grids.

\begin{figure}[!ht]
    \centerline{\includegraphics[width=\columnwidth]{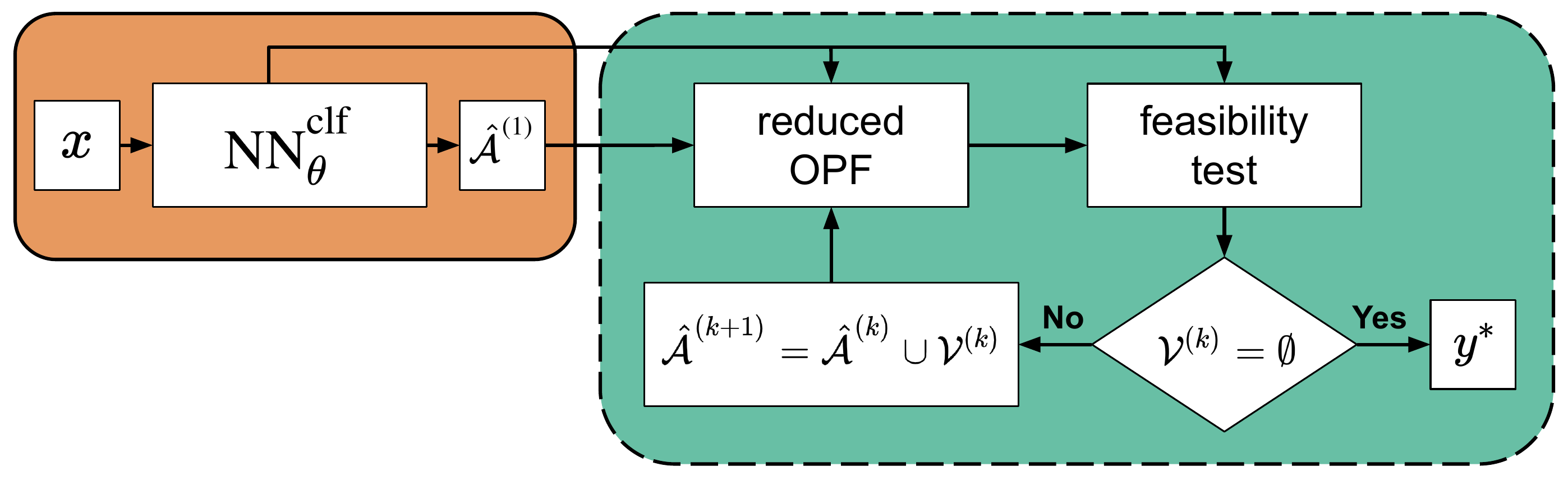}}
    \caption{Flowchart of the iterative feasibility test method (green panel) combined with a NN classifier (orange panel). $\hat{\mathcal{A}}^{(k)}$ and $\mathcal{V}^{(k)}$ are the sets of predicted active and violated inequality constraints at the $k$-th step of the iterative feasibility test, respectively.  For clarity, default arguments of the OPF operator are omitted.}
    \label{fig:reduced_problem}
\end{figure}

To alleviate the security risks associated with imperfect classification, an \textit{iterative feasibility test} can be employed to reinstate violated constraints until convergence, as detailed in \cite{Robson2020} (Figure~\ref{fig:reduced_problem}). Since the reduced OPF is much cheaper relative to the full problem, this approach can in theory be rather efficient.

\subsection{Contributions}
Both the end-to-end and hybrid techniques for machine learning assisted OPF benefit from NN architectures designed to maximize predictive performance. Related works typically employ a range of shallow to deep fully connected neural networks (FCNN). However, convolutional (CNN) \cite{Chen2020} and graph (GNN) \cite{Owerko2020, FalconerMones2020, Falconer2020} neural networks have recently been investigated to exploit assumed locality properties within the respective power grid, i.e. whether the topology of the electricity network influences the correlation between inputs and outputs. 

Building on this set of works, our contributions are as follows: 
\begin{itemize}
    \item We introduce a concise framework for generalizing end-to-end and hybrid methods for machine learning assisted OPF by characterising them as estimators of the corresponding OPF operator or function.
    \item We provide a systematic comparison between the aforementioned NN architectures for both the regression and classification approaches.
    \item We demonstrate the marginal utility of applying CNN and GNN architectures for \emph{fixed topology} problems (i.e. varying grid parameters only for the same topology), hence recommend the application of FCNN models for such problems.
    \item We show that locality properties between grid parameters (features or inputs) and corresponding generator set-points (targets or outputs) -- essential for efficient inductive bias in both CNN and GNN models -- are weak, which explains the moderate performance of these models compared to FCNN.
    \item We also show that a similar weak locality applies between grid parameters and locational marginal prices (LMPs), indicating that the applicability of CNN and GNN architectures would face similar challenges if instead used to predict these derived market signals.
    \item We present a set of \emph{varying topology} problems (i.e. when both grid parameters and network topology are varied), that demonstrate successful utilization of structure based inductive bias through superior predictive performance of GNN models relative to both CNN and FCNN models.
\end{itemize}

It should be noted that, although we address the requirement of accurate predictions for machine learning assisted OPF, feasibility and optimality concerns associated with end-to-end methods, as well as the computational limitation of hybrid methods, remains a challenge for future work.

\section{Methodology}

\subsection{Problem formulation}
This work centers on the fundamental form of OPF, without consideration for unit commitment or security constraints (although machine learning assisted OPF can be readily extended to such cases \cite{Halilbasic2018, Pan2020}).
In general, OPF problems can be expressed using the following concise form of mathematical programming:

\begin{equation}
    \begin{aligned}
        & \min \limits_{y}\ f(x, y) \\
        & \mathrm{s.\ t.} \ \ c_{i}^{\mathrm{E}}(x, y) = 0 \quad i = 1, \dots, n \\
        & \quad \; \; \; \; \; c_{j}^{\mathrm{I}}(x, y) \ge 0 \quad j = 1, \dots, m \\
    \end{aligned}
    \label{opt}
\end{equation}
where $x$ and $y$ are the vectors of grid parameters and optimization variables, respectively, $f(x, y)$ is the objective (or cost) function (parameterized by $x$), which is minimized with respect to $y$ and subject to equality constraints $c_{i}^{\mathrm{E}}(x, y) \in \mathcal{C}^{\mathrm{E}}$ and inequality constraints $c_{j}^{\mathrm{I}}(x, y) \in \mathcal{C}^{\mathrm{I}}$. For convenience, we introduce $\mathcal{C}^{\mathrm{E}}$ and $\mathcal{C}^{\mathrm{I}}$, which denote the sets of equality and inequality constraints with corresponding cardinalities $n = \lvert \mathcal{C}^{\mathrm{E}} \rvert$ and $m = \lvert \mathcal{C}^{\mathrm{I}} \rvert$. For instance, in a simple economic dispatch problem (the focus of this work), $x$ includes the active and reactive power components of loads, $y$ is a vector of voltage magnitudes and active powers of generators and the objective function is a quadratic or piece-wise linear function of the (monotonically increasing) generator cost curves. Equality constraints include the power balance and power flow equations, whilst inequality constraints impose lower and upper bounds on certain quantities.

\subsection{OPF operators and functions}
By formulating the problem in such a manner as $\eqref{opt}$, one can view OPF as an operator, which maps the grid parameters ($x$) to the optimal value of the optimization variables ($y^{*}$) \cite{Zhou2020}. In order to introduce a consistent framework, we extend the operator arguments by the objective ($f$) and constraint functions ($\mathcal{C}^{\mathrm{E}}$ and $\mathcal{C}^{\mathrm{I}}$), as well as by the starting value of the optimization variables ($y^{0}$). The value of $y^{0}$ has a considerable influence of the convergence rate of interior-point methods, and for non-convex formulations with multiple possible local minima, even the found optimum is a function of $y^{0}$. The general form of the OPF operator can be written as\footnote{We note that an even more general form of the operator can be defined when the arguments are mapped to the joint space of the primal and dual variables of the optimization problem: $\Psi: \Omega \to \mathbb{R}^{n_{y} + n_{z}}: \quad \Psi \left( x, y^{0}, f, \mathcal{C}^{\mathrm{E}}, \mathcal{C}^{\mathrm{I}} \right) = (y^{*}, z^{*})$, where $z^{*}$ is the optimal value of the Lagrangian multipliers of the equality and inequality constraints. As locational marginal prices are computed from $z^{*}$, this formalism is useful to construct estimators for learning electricity prices.}:

\begin{equation}
    \Phi: \Omega \to \mathbb{R}^{n_{y}}: \quad \Phi\left( x, y^{0}, f, \mathcal{C}^{\mathrm{E}}, \mathcal{C}^{\mathrm{I}} \right) = y^{*},
    \label{opf-operator}
\end{equation}
where $\Omega$ is an abstract set within which the values of the operator arguments are allowed to change and $n_{y}$ denotes the dimension of the optimization variables. In the simplest case, only the grid parameters vary, whilst most arguments of the OPF operator remain fixed. Accordingly, we introduce a simpler notation, the OPF function, for such cases:

\begin{equation}
    F_{\Phi}: \mathbb{R}^{n_{x}} \to \mathbb{R}^{n_{y}}: \quad F_{\Phi}(x) = y^{*},
    \label{opf-function}
\end{equation}
where $n_{x}$ and $n_{y}$ are the dimensions of the grid parameters and optimization variables, respectively, whilst $\mathcal{F}_{\Phi}$ is used to denote the set of all feasible points, such that $y^{*} \in \mathcal{F}_{\Phi}$. Depending on the grid parameters, the problem may be infeasible: $\mathcal{F}_{\Phi} = \emptyset$.

\subsection{Estimators of OPF operators and functions}
Machine learning assisted OPF methods apply either an estimator operator or function, which both provide a computationally cheap prediction to the optimal point of the OPF based on the grid parameters, i.e. $\hat{\Phi}(x) = \hat{y}^{*}: \| \hat{y}^{*} - y^{*} \| < \varepsilon \ \land \  \mathbb{T}[\hat{\Phi}] \ll \mathbb{T}[\Phi]$ and $\hat{F}_{\Phi}(x) = \hat{y}^{*}: \| \hat{y}^{*} - y^{*} \| < \varepsilon \ \land \ \mathbb{T}[\hat{F}_{\Phi}] \ll \mathbb{T}[F_{\Phi}]$, where $\| \cdot \|$ is an arbitrary norm, $\varepsilon$ is a threshold variable and $\mathbb{T}$ denotes the computational time to obtain the solution.

\subsubsection{End-to-end}
To learn the optimal OPF solution directly from the grid parameters, NNs as regressors can be used, depicted by the following function:

\begin{equation}
    \hat{F}_{\Phi}(x) = \mathrm{NN}_{\theta}^{\mathrm{reg}}(x) = \hat{y}^{*},
\end{equation}
where subscript $\theta$ denotes the NN parameters and the superscript $\mathrm{reg}$ indicates that the NN is used as a regressor. The problem dimensionality can be reduced by predicting only a subset of the optimization variables -- in this case, the remaining state variables can be easily obtained by solving the corresponding power flow problem \cite{Zamzam2019}, given the prediction is a feasible point. Optimal NN parameters can be obtained by minimizing some loss function between the ground-truth $y^*$ and prediction $\hat{y}^{*}$ of some training set. Typically, the squared L2-norm, i.e. mean-squared error (MSE), is used: $\ell(y^*,\hat{y}^*) = \| y^*-\hat{y}^* \|_2^2$. To mitigate violations of certain constraints, a penalty term can be added to this loss function \cite{Fioretto2019}.

\subsubsection{Warm-start}
Warm-start approaches utilize a hybrid model whereby a NN is first parameterized to infer an approximate set-point, $\hat{y}^{0} = \mathrm{NN}_{\theta}^{\mathrm{reg}}(x)$, which is subsequently used to initialize the constrained optimization procedure resulting in the exact solution ($y^{*}$): 

\begin{align}
    \hat{\Phi}^{\mathrm{warm}}(x) & = \Phi \left( x, \hat{y}^{0}, f, \mathcal{C}^{\mathrm{E}}, \mathcal{C}^{\mathrm{I}} \right) \\ & = \Phi \left( x, \mathrm{NN}_{\theta}^{\mathrm{reg}}(x), f, \mathcal{C}^{\mathrm{E}}, \mathcal{C}^{\mathrm{I}} \right) \\ & = y^{*}.
\end{align}

Optimal NN parameters can be obtained by minimizing a similar conventional loss function as in the case of the end-to-end approach. However, significant improvement has been demonstrated by optimizing NN parameters with respect to a (meta-)loss function corresponding directly to the time complexity of the entire pipeline (i.e. including the warm-started OPF) \cite{Jamei2019}: $\ell(\hat{y}^0) = \mathbb{T} \left[\Phi \left( x, \hat{y}^{0}, f, \mathcal{C}^{\mathrm{E}}, \mathcal{C}^{\mathrm{I}} \right)\right]$.

\subsubsection{Reduced problem}
In this hybrid approach, a binary multi-label NN classifier ($\mathrm{NN}_{\theta}^{\mathrm{clf}}$) is used to predict the active set of constraints, and a reduced OPF problem is formulated, which maintains the same objective function as the original full problem:

\begin{align}
    \hat{\Phi}^{\mathrm{red}}(x) & = \Phi \left( x, y^{0}, f, \mathcal{C}^{\mathrm{E}}, \hat{\mathcal{A}} \right) \\ & = \Phi \left( x, y^{0}, f, \mathcal{C}^{\mathrm{E}}, \mathrm{NN}_{\theta}^{\mathrm{clf}}(x) \right) \\ &= \hat{y}^{*},
\end{align}
where $\mathcal{A} \subseteq \mathcal{C}^{\mathrm{I}}$ is the active subset of the inequality constraints and $\hat{\mathcal{A}}$ is the predicted active set. It should also be noted that $\mathcal{C}^{\mathrm{E}} \cup \mathcal{A}$ contains all active constraints defining the specific congestion regime. In the case of a multi-label classifier, the output is a binary vector representing an enumeration of the set of non-trivial constraints, learnt by minimizing the binary cross-entropy (BCE) loss between the ground-truths represented by $\mathcal{A}$ and the predicted binding probabilities of constraints defining $\hat{\mathcal{A}}$: $\ell(\mathcal{A}, \hat{\mathcal{A}}) = -\sum \limits_{j} c_{j} \log \hat{c}_{j} + (1 - c_{j}) \log (1 - \hat{c}_{j})$. The output dimension of the multi-label classifier is reduced by removing trivial constraints (those that are always binding or non-binding in the training set) for training. We note that to formulate the subsequent reduced OPF problem, these constraints need to be reinstated before the iterative feasibility test to construct the complete active set.

Violated constraints omitted from the reduced model are retained using the aforementioned iterative feasibility test to ensure convergence to an optimal point of the full problem.  The computational gain can again be further enhanced via meta-optimization by directly encoding the time complexity into a (meta-)loss function and optimizing the NN weights accordingly \cite{Robson2020}: $\ell(\hat{\mathcal{A}}) = \mathbb{T} \left[\Phi \left( x, y^{0}, f, \mathcal{C}^{\mathrm{E}}, \hat{\mathcal{A}} \right)\right]$.

\subsection{Architectures}
\begin{figure}[!ht]
    \centerline{\includegraphics[width=\columnwidth]{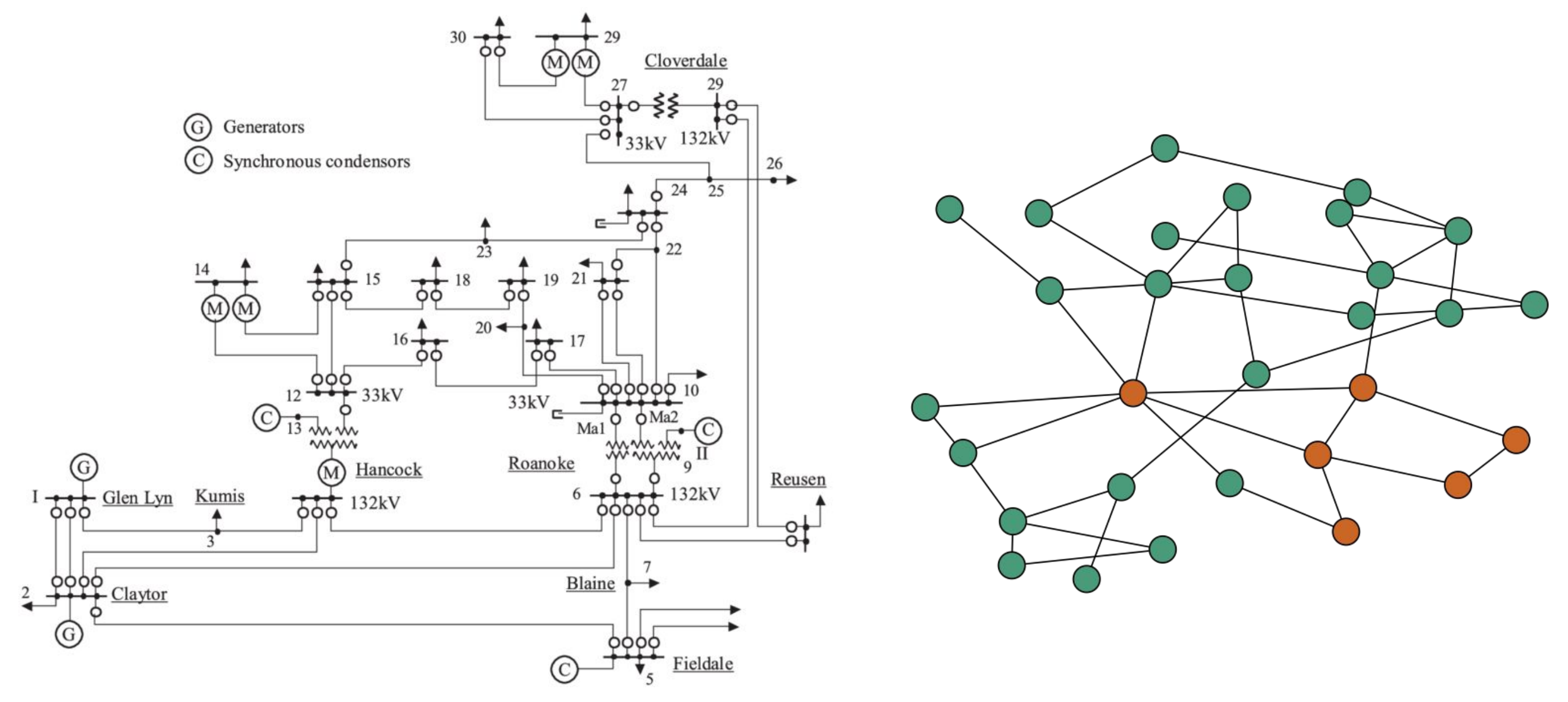}}
    \caption{Schematic diagram \cite{UW2021} (left) and corresponding graphical representation (right) for synthetic grid 30-ieee. Orange and green circles denote generator and load buses, respectively.}
    \label{fig:graph}
\end{figure}

Power grids are complex networks consisting of buses (e.g. generation points, load points etc.) connected by transmission lines, hence can conveniently be depicted as an un-directed graph $\Graph = (\Node, \Edge)$, where $\Node$ and $\Edge \subseteq \Node \times \Node$ denote the sets of nodes and edges (Figure \ref{fig:graph}). Also, $\mathcal{G}$ and $\mathcal{L}$ will denote the sets of generators and loads, respectively.

This formulation motivates the use of NN architectures specifically designed to leverage the spatial dependencies within non-Euclidean data structures, i.e. GNN models -- the hypothesis being that OPF problems exhibit a locality property whereby the network topology influences to correlation between grid parameters and the subsequent solution.

In real power grids, however, a given bus can include multiple generators and loads, which, although can have different power supply and demand, share the bus voltage. To accommodate such characteristics in GNN models straightforwardly, we use a transformed version of the original graph: $\Graph^\prime = (\Node^\prime, \Edge^\prime)$, where each node of the transformed network represents either a single generator or a load (i.e. $|\Node^\prime| = |\Generator| + |\Load|$), and generators and loads belonging to the same bus of the original network are interconnected. With this representation of the grid, generator real power outputs are obtained as individual nodal features, while bus voltage magnitudes are computed as averages of the corresponding individual voltages.

\subsubsection{FCNN}
Fully connected NN models, denoted by $\mathcal{M}^{\mathrm{FCNN}}$, are used here as baseline. Their input domain is equivalent to the raw vector of grid parameters, i.e. active and reactive power components of loads: $x \in \mathbb{R}^{2 |\Load|}$, while the corresponding output vector includes the generators' injected active power and the voltage magnitude at buses comprising at least one generator $(\Node^\textrm{gen} \in \Node)$, i.e. $y \in \mathbb{R}^{|\Generator| + |\Node^\textrm{gen}|}$. Since FCNNs are defined in an un-structured data space, this baseline theoretically lacks sufficient relational inductive bias to efficiently exploit any underlying spatial dependencies -- this information could be learnt implicitly through optimization, but possibly requires a highly flexible model with a large amount of data, thus scaling poorly to large-scale OPF problems \cite{Dehmamy2019}. We investigated two FCNN models using one $(\mathcal{M}^{\textrm{FCNN}}_{\textrm{global-1}})$ and three $(\mathcal{M}^{\textrm{FCNN}}_{\textrm{global-3}})$ hidden layers.

\subsubsection{CNN}
We explore the utility of augmenting the fully connected layers with an antecedent sequence of convolutional and pooling layers $(\mathcal{M}^{\textrm{CNN}}_{\textrm{global-4}})$, designed to extract a spatial hierarchy of latent features, which are subsequently (non-linearly) mapped to the target. A reasonable assumption here is that one can leverage spatial correlations within pseudo-images of the electrical grid using the weighted adjacency matrix. However, convolutions in Euclidean space are dependent upon particular geometric priors, which are not observed in the graph domain (e.g. shift-invariance), hence filters can no longer be node-agnostic and the lack of natural order means operations need to instead be permutation invariant. Nevertheless, we validate this conjecture using CNNs by combining each load constituent of length $|\Node^\prime|$ into a 3-dimensional tensor, i.e. $x \in \mathbb{R}^{2 \times |\Node^\prime| \times |\Node^\prime|}$.

\subsubsection{GNN}
We analyze several GNN architectures whereby the weighted adjacency matrix is used to extract latent features by propagating information across neighbouring nodes irrespective of the input sequence \cite{Zhou2018}. Such propagation is achieved using graph convolutions, which can be broadly categorized as either spectral or spatial filtering \cite{Zonghan2021}.

Spectral filtering adopts methods from graph signal processing: operations occur in the Fourier domain whereby input signals are passed through parameterized functions of the normalized graph Laplacian, thereby exploiting its positive-semidefinite property. Given this procedure has $\mathcal{O}(|\Node^\prime|^3)$ time complexity, we investigate four spectral layers designed to reduce computational costs by avoiding full eigendecomposition of the Laplacian: (1) \textit{ChebConv} $(\mathcal{M}^{\textrm{CHC}})$, which uses approximate filters derived from Chebyshev polynomials of the eigenvalues up to the $K$-th order \cite{Kipf2017}; (2) \textit{GCNConv} $(\mathcal{M}^{\textrm{GCN}})$, which constrains the layer-wise convolution to first-order neighbours ($K=1$), lessening overfitting to particular localities \cite{Defferrard2017}; (3) \textit{GraphConv} $(\mathcal{M}^{\textrm{GC}})$, which is analogous to \textit{GCNConv} except adapting a discrete weight matrix for self-connections \cite{Weisfeiler2020}; and (4) \textit{GATConv} $(\mathcal{M}^{\textrm{GAT}})$, which extends the message passing framework of \textit{GCNConv} by assigning each edge with relative importance through attention coefficients \cite{Velickovic2018}.

By contrast, spatial graph convolutions (a non-Euclidean generalization of the convolution operation  found in CNNs) are performed directly in the graph domain, reducing the computational complexity whilst minimizing loss of structural information -- a byproduct of reducing to embedded space \cite{Wu2020}. We investigate \textit{SplineConv} $(\mathcal{M}^{\textrm{SC}})$ \cite{Fey2018} which, for a given node, computes a linear combination of its features together with those of its $K$-th order neighbours, weighted by a kernel function -- the product of parameterized B-spline basis functions. The local support property of B-splines reduces the number of parameters, enhancing the computational efficiency of the operator. Note that all GNN models are named in accordance with the \texttt{PyTorch Geometric} library \cite{Fey2019}.

Finally, we note that due to the lack of connectivity information of the grid, conventional FCNN (and CNN) architectures typically fail to adapt efficiently to power system restructuring. In order to obtain sufficient performance with alternative grid topologies (i.e. contingency cases), these models need to be re-trained with appropriate training data. In contrast, GNNs compute localized convolutions in a manner such that the number of weights remains independent of the topology of the network making these models capable to train and predict on samples having different topologies \cite{Zonghan2021}.

\subsection{Technical details}
\subsubsection{Samples}
To span multiple grid sizes, we built test cases using several synthetic grids from the Power Grid Library \cite{Babaeinejadsarookolaee2019} ranging from 24 -- 2853 buses. To maintain validity of the constructed data sets whilst ensuring a thorough exploration of congestion regimes, we generated 10k (feasible) fixed topology samples for each synthetic grid by re-scaling each active and reactive load component (relative to nominal values) by factors independently drawn from a uniform distribution, $\mathcal{U}(0.8, 1.2)$. To investigate performance of the different NN architectures with varying topology, we also generated 10k (feasible) samples subject to \nmo{} line contingency. For each sample, active and reactive load components were re-scaled as before and a single transmission line was randomly removed from the original grid topology. OPF solutions were obtained using \texttt{PowerModels.jl} \cite{Coffrin2018} (an OPF package written in Julia \cite{Bezanson2012}) in combination with the \texttt{IPOPT} solver \cite{Wachter2006}.

\subsubsection{Neural Networks}
Our model with the largest number of parameters was the three hidden layer fully connected model ($\mathcal{M}^{\textrm{FCNN}}_{\textrm{global-3}}$) that also served as the baseline. The size of each hidden layer was computed through a linear interpolation between the corresponding input and output sizes. 

In the case of CNN, each model was constructed using \(3\times 1\) kernels, 1-dimensional max-pooling layers, zero-padding and a stride length of 1. 

For GNN models, we investigated three architecture types:
(1) the first type included two convolutional layers followed by a fully connected readout layer making the original local structure non-local ($\mathcal{M}^{\textrm{GNN}}_{\textrm{global-3}}$);
(2) in the second type, only three convolutional layers were present, simply treating the features available locally at each node as the output ($\mathcal{M}^{\textrm{GNN}}_{\textrm{local-3}}$); and lastly (3) the third type was again a global one extending the above local type with a fully connected readout layer ($\mathcal{M}^{\textrm{GNN}}_{\textrm{global-4}}$).
While corresponding $\mathcal{M}^{\textrm{GNN}}_{\textrm{global-3}}$ and $\mathcal{M}^{\textrm{GNN}}_{\textrm{local-3}}$ models were constructed to have an approximately equal number of parameters (details discussed below), $\mathcal{M}^{\textrm{GNN}}_{\textrm{global-4}}$ models had a significantly larger number of parameters due to the additional readout layer.
For $\mathcal{M}^{\textrm{CHC}}$ and $\mathcal{M}^{\textrm{SC}}$ models, the hyperparameter $K$ was set to 4.

Since our aim was to compare the predictive performance of models with and without topology based inductive bias, the single-layer FCNN, CNN and several GNN architectures were constructed to have a similar number of parameters for each synthetic grid.
This required scaling the number of channels of the hidden layers of some architectures according to both the grid size ($\sigma_{s}$) and the model type ($\sigma_{m}$).
We applied a simple grid search in order to obtain the optimal number of layers, as well as the values of parameters $\sigma_{s}$ and $\sigma_{m}$.
The actual number of channels used for the CNN and GNN models is presented in Table~\ref{tab:layers}.

Edge weights ($e_{ij}$) of the GNN architectures were modeled as a function of transmission line impedance, $Z_{ij}$, between the $i$-th and $j$-th bus. Specifically, we used the following general expression between connected buses $i$ and $j$:

\begin{equation}
    e_{ij} = \exp(-k \log|Z_{ij}|),
    \label{edge-weights}
\end{equation}
where $k$ is a hyperparameter.
Note that $k = 0$ leads to the application of the simple binary adjacency matrix, while in the case of $k = 1$ the absolute value of the corresponding element of the nodal admittance matrix is used.

For each grid, the generated 10k samples were split into training, validation and test sets with a ratio of 80:10:10. In all cases, the ADAM \cite{Diederik2014} optimizer was applied (with default parameters $\beta_1$ = 0.9 and $\beta_2$ = 0.999 and learning-rate $\eta=10^{-4}$) using an early stopping with a patience of 20 determined on the validation set. Mini-batch size of 100 was applied and hidden layers were equipped with BatchNorm \cite{Ioffe2015} and a ReLU \cite{Xu2015} activation function was used.
For each model, statistics (mean and two-sided 95\% confidence interval) of the predictive performance were computed using 10 independent runs. 

Models were implemented in Python 3.0 using \texttt{PyTorch} \cite{Paszke2019} and \texttt{PyTorch Geometric} \cite{Fey2019} libraries.
Experiments were carried out on NVIDIA Tesla M60 GPUs. In order to facilitate research reproducibility in the field, we have made the generated samples, as well as the code our work is based upon, publicly available at \url{https://github.com/tdfalc/MLOPF.jl}.

\begin{table}[!h]
    \caption{Number of channels used for CNN and GNN architectures. $\sigma_{s}$ and $\sigma_{m}$ are the grid size and model type based scaling factors. $n_{n}$ denotes the number of nodes of the transformed network and $n_{y}$ is the number of output variables.\\\hspace{\textwidth}
    $\sigma_{s} = \begin{cases} 1 & \mathrm{if\ |\Node| \le 73} \\ 2 & \mathrm{if\ |\Node| > 73} \\ \end{cases}$ \ \ 
    $\sigma_{m} = \begin{cases} 1 & \mathrm{if \ \mathcal{M} = \mathcal{M}^{\textrm{GCN}} \ or\ \mathcal{M}^{\textrm{GAT}}} \\ 0.5 & \mathrm{if \ \mathcal{M} = \mathcal{M}^{\textrm{CHC}}} \end{cases}$}
    \label{tab:layers}
    \centering
    \begin{tabular}{lllll}
        \toprule
        GNN layer &
        $\mathcal{M}^{\textrm{CNN}}_{\textrm{global-4}}$ & $\mathcal{M}^{\textrm{GNN}}_{\textrm{global-3}}$ & $\mathcal{M}^{\textrm{GNN}}_{\textrm{local-3}}$ & $\mathcal{M}^{\textrm{GNN}}_{\textrm{global-4}}$ \\
        \midrule
        1. & 4 & 8$\sigma_{s}$ & 8 & 8 \\
        2. & 8 & 16$\sigma_{s}$ & $n_{n}\sigma_{m}$ & $n_{n}\sigma_{m}$ \\
        3. & 16 & --- & $n_{y}\sigma_{s}\sigma_{m}$ & $n_{y}\sigma_{s}\sigma_{m}$ \\
        \midrule
        Readout layer & yes & yes & no & yes \\
        \bottomrule
    \end{tabular}
\end{table}

\section{Numerical Results}

\subsection{Computational performance of prediction}
The fundamental motivation for using NN models to predict OPF solutions is their superior (online) computational performance compared to directly solving the corresponding AC-OPF problems.
In Table~\ref{tab:pred_train_times_reg_global}, we compared the average computational times of obtaining \textit{exact} AC-OPF solutions using the \texttt{IPOPT} solver against inferring approximate solutions using various NN architectures.
It is evident that, for all investigated systems, the computational time of the NN models is several orders of magnitude smaller than that of solving AC-OPF with conventional methods (note that in Table~\ref{tab:pred_train_times_reg_global}, solve times of AC-OPF refer to a single sample, while prediction times of NN models refer to 1000 samples).
Constrained optimization problems were solved on CPU (Intel Xeon E5-2686 v4, 2.3 GHz), while for the NN predictions we could utilize GPU (NVIDIA Tesla M60).

However, as discussed previously, comparing these computational times alone can be misleading: NN predictions are not necessarily optimal or even feasible.
There have been several attempts to obtain feasible and possibly optimal estimates of OPF solutions (for instance by using hybrid approaches \cite{Zamzam2019, Pan2020} or introducing penalty terms of constraint violations in the loss function \cite{Fioretto2019}).
For all approaches, improving the quality of the predictive performance is fundamental.
One apparent way is to increase the training data size significantly.
In the following, we investigate the applicability of a more economical approach by using appropriate inductive bias in NN models.

\begin{table*}[!ht]
\small
\caption{Prediction time statistics (mean and two-sided 95\% confidence intervals) for global regression models}
\label{tab:pred_train_times_reg_global}
\def\na{---}
\centering
    \resizebox{\textwidth}{!}{
    \begin{tabular}{lr|rrrrrrr}
    \toprule
    \multirow{2}{*}{Case} & Solve time (ms) & \multicolumn{7}{c}{Prediction time per 1000 samples (ms)} \\
    \cmidrule(r){2-9}
    & 
    $\textrm{AC-OPF} \ \mathrm{(IPOPT)}$ & $\mathcal{M}^{\textrm{FCNN}}_{\textrm{global-1}}$ &  $\mathcal{M}^{\textrm{CNN}}_{\textrm{global-4}}$ & $\mathcal{M}^{\textrm{GCN}}_{\textrm{global-3}}$ & $\mathcal{M}^{\textrm{CHC}}_{\textrm{global-3}}$ & $\mathcal{M}^{\textrm{SC}}_{\textrm{global-3}}$ & $\mathcal{M}^{\textrm{GC}}_{\textrm{global-3}}$ & $\mathcal{M}^{\textrm{GAT}}_{\textrm{global-3}}$ \\
    \midrule
    $\textrm{24-ieee-rts}$ & $85.41 \pm 1.04$ & $10.86 \pm 0.13$ & $20.19 \pm 0.11$ & $191.75 \pm 0.37$ & $251.64 \pm 2.99$ & $196.64 \pm 3.83$ & $191.72 \pm 0.75$ & $236.52 \pm 42.51$ \\
    $\textrm{30-ieee}$ & $75.33 \pm 0.63$ & $10.58 \pm 0.08$ & $20.07 \pm 0.11$ & $194.45 \pm 0.48$ & $254.36 \pm 2.16$ & $197.59 \pm 3.95$ & $193.64 \pm 0.88$ & $237.33 \pm 42.54$ \\
    $\textrm{39-epri}$ & $147.47 \pm 1.28$ & $11.31 \pm 0.14$ & $21.47 \pm 0.14$ & $203.67 \pm 1.99$ & $269.31 \pm 1.91$ & $208.38 \pm 4.43$ & $204.75 \pm 2.46$ & $248.08 \pm 43.68$ \\
    $\textrm{57-ieee}$ & $125.24 \pm 1.16$ & $11.36 \pm 0.06$ & $21.06 \pm 0.25$ & $196.32 \pm 0.27$ & $257.18 \pm 3.76$ & $200.19 \pm 4.55$ & $196.12 \pm 0.91$ & $238.82 \pm 41.02$ \\
    $\textrm{73-ieee-rts}$ & $304.64 \pm 1.32$ & $13.25 \pm 0.18$ & $23.09 \pm 0.41$ & $216.67 \pm 3.93$ & $285.72 \pm 7.28$ & $220.83 \pm 7.14$ & $214.43 \pm 3.25$ & $260.34 \pm 39.12$ \\
    $\textrm{118-ieee}$ & $481.39 \pm 2.68$ & $12.59 \pm 0.08$ & $23.64 \pm 1.94$ & $200.02 \pm 0.31$ & $267.59 \pm 3.88$ & $203.14 \pm 3.68$ & $198.96 \pm 0.29$ & $245.38 \pm 39.21$ \\
    $\textrm{162-ieee-dtc}$ & $815.66 \pm 6.27$ & $13.81 \pm 0.17$ & $25.62 \pm 2.39$ & $207.86 \pm 3.52$ & $285.46 \pm 7.57$ & $215.16 \pm 6.97$ & $205.53 \pm 3.72$ & $261.93 \pm 44.27$ \\
    $\textrm{300-ieee}$ & $1467.43 \pm 9.47$ & $16.36 \pm 0.08$ & $28.04 \pm 2.03$ & $206.19 \pm 0.74$ & $301.14 \pm 4.46$ & $240.13 \pm 3.28$ & $203.19 \pm 0.89$ & $279.32 \pm 42.01$ \\
    $\textrm{588-sdet}$ & $2826.53 \pm 51.2$ & $22.03 \pm 0.24$ & $34.43 \pm 2.36$ & $240.94 \pm 1.04$ & $422.67 \pm 3.56$ & $363.13 \pm 5.57$ & $235.18 \pm 0.64$ & $354.07 \pm 41.78$ \\
    $\textrm{1354-pegase}$ & $10814.92 \pm 29.6$ & $36.04 \pm 0.63$ & $52.15 \pm 8.89$ & $390.56 \pm 6.59$ & $751.89 \pm 9.86$ & $676.29 \pm 7.58$ & $413.22 \pm 4.71$ & $520.68 \pm 79.02$ \\
    $\textrm{2853-sdet}$ & $34136.73 \pm 99.1$ & $76.54 \pm 1.55$ & $98.42 \pm 2.42$ & $1092.19 \pm 5.79$ & $1729.84 \pm 8.66$ & $1520.66 \pm 9.32$ & $1116.61 \pm 9.94$ & $1246.24 \pm 39.39$ \\
    \bottomrule
    \end{tabular}
    }
\end{table*}

\subsection{Fixed topology}
We begin our analysis by investigating the predictive performance of NN models trained (and tested) using data derived from power grids with a fixed topology. In these experiments, only the grid parameters were varied within the datasets, while all the grid connections were the same among the samples. 
In this setup, FCNN and CNN architectures are functions of the grid parameters only, i.e. for regression and classification approaches we have $\mathrm{NN}_{\theta}^{\mathrm{reg}}(x_{i}) = \hat{y}_{i}^{*}$ and $\mathrm{NN}_{\theta}^{\mathrm{clf}}(x_{i}) = \hat{\mathcal{A}}_{i}$, where $x_{i}$ is the grid parameter vector of the $i$-th sample.
For GNN models, besides the grid parameters, the grid topology is also passed: $\mathrm{NN}_{\theta}^{\mathrm{reg}}(x_{i}, \mathbb{G}) = \hat{y}_{i}^{*}$ and $\mathrm{NN}_{\theta}^{\mathrm{clf}}(x_{i}, \mathbb{G}) = \hat{\mathcal{A}}_{i}$, where $\mathbb{G}$ represents the (fixed) grid topology with corresponding edge weights.

\subsubsection{Regression}
For each grid, Table~\ref{tab:error_reg_global} summarizes the MSE statistics for regression model architectures that encode the targets as global variables.
The first column includes the results of our baseline $\mathcal{M}^{\textrm{FCNN}}_{\textrm{global-3}}$ model, which has the largest number of parameters (Table~\ref{tab:params_reg_global}).
In the presence of appropriate locality attributes, CNN and GNN models are expected to provide a comparable performance to $\mathcal{M}^{\textrm{FCNN}}_{\textrm{global-3}}$ with a significantly smaller amount of parameters due to their topology based inductive bias.

\begin{table*}[!ht]
\small
\caption{MSE statistics (mean and two-sided 95\% confidence intervals) of the test sets for global regression models (fixed topology)}
\label{tab:error_reg_global}
\def\na{---}
\centering
    \resizebox{\textwidth}{!}{
    \begin{tabular}{lr|rrrrrrr}
    \toprule
    \multirow{2}{*}{Case} & \multicolumn{8}{c}{MSE ($\times 10^{-3}$)} \\
    \cmidrule(r){2-9}
    & 
    $\mathcal{M}^{\textrm{FCNN}}_{\textrm{global-3}}$ & $\mathcal{M}^{\textrm{FCNN}}_{\textrm{global-1}}$ & $\mathcal{M}^{\textrm{CNN}}_{\textrm{global-4}}$ & $\mathcal{M}^{\textrm{GCN}}_{\textrm{global-3}}$ & $\mathcal{M}^{\textrm{CHC}}_{\textrm{global-3}}$ & $\mathcal{M}^{\textrm{SC}}_{\textrm{global-3}}$ & $\mathcal{M}^{\textrm{GC}}_{\textrm{global-3}}$ & $\mathcal{M}^{\textrm{GAT}}_{\textrm{global-3}}$ \\
    \midrule
    $\textrm{24-ieee-rts}$ & $0.18 \pm 0.02$ & $0.94 \pm 0.04$ & $1.55 \pm 0.21$ & $2.65 \pm 0.13$ & $\textbf{0.70} \boldsymbol{\pm} \textbf{0.04}$ & $1.10 \pm 0.12$ & $1.04 \pm 0.06$ & $2.76 \pm 0.19$ \\
    $\textrm{30-ieee}$ & $0.05 \pm 0.01$ & $\textbf{0.03}     \boldsymbol{\pm} \textbf{0.01}$ & $0.62 \pm 0.22$ & $3.25 \pm 0.82$ & $0.09 \pm 0.01$ & $0.27 \pm 0.08$ & $0.26 \pm 0.12$ & $3.06 \pm 0.33$ \\
    $\textrm{39-epri}$ & $0.89 \pm 0.10$ & $3.16 \pm 0.09$ & $7.01 \pm 0.09$ & $4.30 \pm 0.23$ & $\textbf{2.38} \boldsymbol{\pm} \textbf{0.10}$ & $3.00 \pm 0.09$ & $2.74 \pm 0.13$ & $4.72 \pm 0.35$ \\
    $\textrm{57-ieee}$ & $0.52 \pm 0.11$ & $1.62 \pm 0.15$ & $\textbf{1.22} \boldsymbol{\pm} \textbf{0.10}$ & $2.18 \pm 0.13$ & $1.28 \pm 0.14$ & $1.64 \pm 0.14$ & $1.59 \pm 0.14$ & $2.28 \pm 0.13$ \\
    $\textrm{73-ieee-rts}$ & $0.21 \pm 0.07$ & $0.69 \pm 0.02$ & $1.06 \pm 0.13$ & $1.59 \pm 0.11$ & $\textbf{0.65} \boldsymbol{\pm} \textbf{0.05}$ & $0.85 \pm 0.11$ & $0.85 \pm 0.07$ & $1.85 \pm 0.21$ \\
    $\textrm{118-ieee}$ & $0.39 \pm 0.03$ & $1.28 \pm 0.07$ & $3.68 \pm 0.75$ & $2.39 \pm 0.12$ & $\textbf{1.23} \boldsymbol{\pm} \textbf{0.07}$ & $1.24 \pm 0.07$ & $1.27 \pm 0.13$ & $2.50 \pm 0.10$ \\
    $\textrm{162-ieee-dtc}$ & $2.61 \pm 0.10$ & $3.19 \pm 0.08$ & $3.28 \pm 0.15$ & $4.77 \pm 0.21$ & $3.08 \pm 0.10$ & $\textbf{2.90} \boldsymbol{\pm} \textbf{0.11}$ & $3.04 \pm 0.10$ & $4.87 \pm 0.23$ \\
    $\textrm{300-ieee}$ & $2.06 \pm 0.06$ & $2.86 \pm 0.05$ & $3.95 \pm 0.22$ & $3.24 \pm 0.09$ & $2.42 \pm 0.04$ & $2.47 \pm 0.20$ & $\textbf{2.39} \boldsymbol{\pm} \textbf{0.06}$ & $3.56 \pm 0.19$ \\
    $\textrm{588-sdet}$ & $2.56 \pm 0.06$ & $3.12 \pm 0.05$ & $4.10 \pm 0.20$ & $4.62 \pm 0.36$ & $3.25 \pm 0.07$ & $\textbf{3.00} \boldsymbol{\pm} \textbf{0.06}$ & $3.05 \pm 0.05$ & $5.07 \pm 0.30$ \\
    $\textrm{1354-pegase}$ & $0.83 \pm 0.12$ & $\textbf{1.30} \boldsymbol{\pm} \textbf{0.09}$ & $2.78 \pm 0.23$ & $2.16 \pm 0.17$ & $1.43 \pm 0.09$ & $1.35 \pm 0.10$ & $1.35 \pm 0.12$ & $2.51 \pm 0.15$ \\
    $\textrm{2853-sdet}$ & $5.99 \pm 0.16$ & $\textbf{6.87} \boldsymbol{\pm} \textbf{0.05}$ & $15.71 \pm 0.93$ & $10.15 \pm 0.58$ & $9.70 \pm 0.33$ & $8.64 \pm 0.29$ & $8.49 \pm 0.41$ & $11.01 \pm 0.46$ \\
    \bottomrule
    \end{tabular}
   }
\end{table*}

In order to investigate the predictive performance with and without topological information, we first constructed global FCNN ($\mathcal{M}^{\textrm{FCNN}}_{\textrm{global-1}}$), CNN ($\mathcal{M}^{\textrm{CNN}}_{\textrm{global-4}}$) and GNN ($\mathcal{M}^{\textrm{GNN}}_{\textrm{global-3}}$) models in a manner such that they have a similar number of parameters for each grid (Table~\ref{tab:params_reg_global}).

\begin{table*}[!ht]
\small
\caption{Number of parameters for global regression models (fixed and varying topology)}
\label{tab:params_reg_global}
\def\na{---}
\centering
    %\resizebox{\columnwidth}{!}{
    \begin{tabular}{lr|rrrrrrr}
    \toprule
    \multirow{2}{*}{Case} & \multicolumn{8}{c}{\# of parameters} \\ 
    \cmidrule(r){2-9}
    & $\mathcal{M}^{\textrm{FCNN}}_{\textrm{global-3}}$ & $\mathcal{M}^{\textrm{FCNN}}_{\textrm{global-1}}$ & $\mathcal{M}^{\textrm{CNN}}_{\textrm{global-4}}$ & $\mathcal{M}^{\textrm{GCN}}_{\textrm{global-3}}$ & $\mathcal{M}^{\textrm{CHC}}_{\textrm{global-3}}$ & $\mathcal{M}^{\textrm{SC}}_{\textrm{global-3}}$ & $\mathcal{M}^{\textrm{GC}}_{\textrm{global-3}}$ & $\mathcal{M}^{\textrm{GAT}}_{\textrm{global-3}}$ \\
    \midrule
    $\textrm{24-ieee-rts}$ & $6575$ & $2156$ & $1336$ & $2303$ & $2783$ & $2943$ & $2463$ & $2353$ \\
    $\textrm{30-ieee}$ & $4436$ & $732$ & $984$ & $607$ & $1087$ & $1247$ & $767$ & $657$ \\
    $\textrm{39-epri}$ & $7877$ & $1580$ & $1568$ & $1035$ & $1515$ & $1675$ & $1195$ & $1085$ \\
    $\textrm{57-ieee}$ & $13933$ & $1610$ & $1722$ & $1047$ & $1527$ & $1687$ & $1207$ & $1097$ \\
    $\textrm{73-ieee-rts}$ & $58677$ & $19404$ & $15504$ & $18715$ & $19195$ & $19355$ & $18875$ & $18765$ \\
    $\textrm{118-ieee}$ & $91835$ & $25596$ & $23160$ & $26354$ & $28178$ & $28786$ & $26962$ & $26454$ \\
    $\textrm{162-ieee-dtc}$ & $104396$ & $7800$ & $7524$ & $8558$ & $10382$ & $10990$ & $9166$ & $8658$ \\
    $\textrm{300-ieee}$ & $440480$ & $82938$ & $78006$ & $83696$ & $85520$ & $86128$ & $84304$ & $83796$ \\
    $\textrm{588-sdet}$ & $1512583$ & $207152$ & $200700$ & $212838$ & $214662$ & $215270$ & $213446$ & $212938$ \\
    $\textrm{1354-pegase}$ & $8486627$ & $1408680$ & $1390548$ & $1409438$ & $1411262$ & $1411870$ & $1410046$ & $1409538$ \\
    $\textrm{2853-sdet}$ & $42568525$ & $9233926$ & $9166558$ & $9299404$ & $9301228$ & $9301836$ & $9300012$ & $9299504$ \\
    \bottomrule
    \end{tabular}
    %}
\end{table*}

In general, the regression performance of the investigated models (including the baseline) has a week correlation with the system size.
This indicates that other factors, for instance the actual number of active sets, can also play an important role (as observed previously in~\cite{Robson2020}).

Comparing the CNN and GNN models, we found that in most of the cases, GNN models outperform the CNN model.
An interesting exception is case 57-ieee, where the CNN model appeared to perform best.
However, we rather consider this as an anomalous case, where the reduced error could be attributed to the coincidental unearthing of structural information within the receptive fields when convolving over the pseudo-image of the grid.

Although GCN is the simplest GNN model we investigated, in general it performs similarly to the more sophisticated GAT model.
Whilst CHC and SC models have similar performance, computational efficiencies with respect to the training times of CHC (Table~\ref{tab:train_times_reg_global}) allude to a better scaling to larger grids.

\begin{table*}[!ht]
\small
\caption{Training time statistics (mean and two-sided 95\% confidence intervals) for global regression models}
\label{tab:train_times_reg_global}
\def\na{---}
\centering
    \resizebox{\textwidth}{!}{
    \begin{tabular}{lr|rrrrrrr}
    \toprule
    \multirow{2}{*}{Case} & \multicolumn{8}{c}{Training time ($\times 10^{2}$ s)} \\
    \cmidrule(r){2-9}
    & 
    $\mathcal{M}^{\textrm{FCNN}}_{\textrm{global-3}}$ & $\mathcal{M}^{\textrm{FCNN}}_{\textrm{global-1}}$ & $\mathcal{M}^{\textrm{CNN}}_{\textrm{global-4}}$ & $\mathcal{M}^{\textrm{GCN}}_{\textrm{global-3}}$ & $\mathcal{M}^{\textrm{CHC}}_{\textrm{global-3}}$ & $\mathcal{M}^{\textrm{SC}}_{\textrm{global-3}}$ & $\mathcal{M}^{\textrm{GC}}_{\textrm{global-3}}$ & $\mathcal{M}^{\textrm{GAT}}_{\textrm{global-3}}$ \\
    \midrule
    $\textrm{24-ieee-rts}$ & $0.75 \pm 0.14$ & $3.40 \pm 0.47$ & $\textbf{1.60} \boldsymbol{\pm} \textbf{0.31}$ & $12.06 \pm 1.76$ & $20.78 \pm 1.97$ & $10.80 \pm 1.87$ & $11.81 \pm 1.67$ & $15.56 \pm 2.64$ \\
    $\textrm{30-ieee}$ & $0.57 \pm 0.05$ & $\textbf{0.58} \boldsymbol{\pm} \textbf{0.04}$ & $1.07 \pm 0.22$ & $9.75 \pm 3.03$ & $16.04 \pm 1.96$ & $9.22 \pm 1.30$ & $14.06 \pm 3.55$ & $22.30 \pm 6.16$ \\
    $\textrm{39-epri}$ & $0.58 \pm 0.10$ & $\textbf{0.82} \boldsymbol{\pm} \textbf{0.05}$ & $0.83 \pm 0.17$ & $12.23 \pm 1.70$ & $16.36 \pm 2.90$ & $8.69 \pm 1.17$ & $9.70 \pm 0.74$ & $15.66 \pm 3.34$ \\
    $\textrm{57-ieee}$ & $0.33 \pm 0.08$ & $\textbf{0.67} \boldsymbol{\pm} \textbf{0.03}$ & $1.13 \pm 0.17$ & $12.73 \pm 1.83$ & $12.39 \pm 2.99$ & $9.20 \pm 2.33$ & $11.93 \pm 2.10$ & $13.69 \pm 2.22$ \\
    $\textrm{73-ieee-rts}$ & $0.83 \pm 0.15$ & $2.79 \pm 0.12$ & $\textbf{1.64} \boldsymbol{\pm} \textbf{0.21}$ & $12.36 \pm 1.82$ & $19.13 \pm 2.19$ & $10.07 \pm 1.49$ & $12.29 \pm 1.92$ & $16.36 \pm 2.53$ \\
    $\textrm{118-ieee}$ & $0.43 \pm 0.09$ & $1.98 \pm 0.18$ & $\textbf{1.66} \boldsymbol{\pm} \textbf{0.28}$ & $17.80 \pm 2.44$ & $8.25 \pm 0.96$ & $7.10 \pm 0.77$ & $5.73 \pm 0.53$ & $20.62 \pm 1.99$ \\
    $\textrm{162-ieee-dtc}$ & $0.28 \pm 0.04$ & $1.32 \pm 0.17$ & $\textbf{1.08} \boldsymbol{\pm} \textbf{0.23}$ & $14.13 \pm 2.56$ & $6.45 \pm 0.83$ & $8.49 \pm 1.69$ & $7.44 \pm 1.29$ & $12.19 \pm 1.89$ \\
    $\textrm{300-ieee}$ & $0.33 \pm 0.02$ & $\textbf{0.64} \boldsymbol{\pm} \textbf{0.05}$ & $1.70 \pm 0.27$ & $14.74 \pm 1.94$ & $11.87 \pm 1.01$ & $13.25 \pm 1.63$ & $8.43 \pm 1.23$ & $16.91 \pm 5.28$ \\
    $\textrm{588-sdet}$ & $0.65 \pm 0.15$ & $\textbf{0.58} \boldsymbol{\pm} \textbf{0.05}$ & $1.84 \pm 0.40$ & $23.74 \pm 6.36$ & $11.24 \pm 1.26$ & $15.54 \pm 3.02$ & $10.72 \pm 1.63$ & $22.61 \pm 4.16$ \\
    $\textrm{1354-pegase}$ & $1.81 \pm 0.22$ & $\textbf{1.13} \boldsymbol{\pm} \textbf{0.11}$ & $1.52 \pm 0.43$ & $18.07 \pm 3.34$ & $22.55 \pm 1.46$ & $26.74 \pm 5.08$ & $13.74 \pm 1.77$ & $21.54 \pm 2.84$ \\
    $\textrm{2853-sdet}$ & $9.54 \pm 0.44$ & $1.37 \pm 0.05$ & $\textbf{0.54} \boldsymbol{\pm} \textbf{0.02}$ & $14.72 \pm 1.00$ & $16.93 \pm 0.88$ & $24.35 \pm 2.19$ & $14.38 \pm 1.24$ & $17.29 \pm 3.29$ \\
    \bottomrule
    \end{tabular}
    }
\end{table*}

The most striking observation is that the single-layer FCNN model exhibits exceedingly comparable performance to the best GNN models.
For several cases, the difference between the average MSE values of the best GNN model and the single-layer model is not statistically significant and for the two largest grids, FCNN even outperforms all GNN models.
It is also worth mentioning that $\mathcal{M}^{\textrm{FCNN}}_{\textrm{global-1}}$ has at least one order of magnitude shorter training times than the global GNN models (Table~\ref{tab:train_times_reg_global}).
For many cases, the significantly larger $\mathcal{M}^{\textrm{FCNN}}_{\textrm{global-3}}$ model had an even shorter training time than $\mathcal{M}^{\textrm{FCNN}}_{\textrm{global-1}}$ due to the faster convergence.

The moderate performance of the global GNN models could be a result of the readout layer, which simply induces noise by arbitrarily mixing signals of nodes further away in the system. 
To investigate this possibility, we performed a set of experiments up to grid size of 588, this time with local architectures for the GCN, CHC and GAT models (left three columns of Table~\ref{tab:error_reg_local}). Interestingly, although the number of parameters of these local models is comparable to that of the global models (Table~\ref{tab:params_reg_local}), the observed performance of each of the three GNN models is considerably worse.
This suggests that the main contribution to the predictive capacity actually stems from the readout layer and also indicates a potential lack of locality properties.

To further validate the above arguments, we investigated the effect of extending the local models with a readout layer, i.e. converting the local regression models to their global counterparts.
We found that using the readout layer significantly improved the predictive performance for all cases (right three columns of Table~\ref{tab:error_reg_local}).

\begin{table*}[!ht]
\small
\caption{MSE statistics (mean and two-sided 95\% confidence intervals) of the test sets for local and extended global regression GNN models (fixed topology)}
\label{tab:error_reg_local}
\def\na{---}
\centering
    %\resizebox{\textwidth}{!}{
    \begin{tabular}{lrrr|rrr}
    \toprule
    \multirow{2}{*}{Case} & \multicolumn{6}{c}{MSE ($\times 10^{-3}$)} \\
    \cmidrule(r){2-7}
    & $\mathcal{M}^{\textrm{GCN}}_{\textrm{local-3}}$ & $\mathcal{M}^{\textrm{CHC}}_{\textrm{local-3}}$ & $\mathcal{M}^{\textrm{GAT}}_{\textrm{local-3}}$ & $\mathcal{M}^{\textrm{GCN}}_{\textrm{global-4}}$ & $\mathcal{M}^{\textrm{CHC}}_{\textrm{global-4}}$ & $\mathcal{M}^{\textrm{GAT}}_{\textrm{global-4}}$ \\
    \midrule
    $\textrm{24-ieee-rts}$ & $73.93 \pm 8.46$ & $\textbf{27.03} \boldsymbol{\pm} \textbf{0.36}$ & $63.69 \pm 9.76$ & $2.63 \pm 0.12$ & $\textbf{0.50} \boldsymbol{\pm} \textbf{0.04}$ & $2.48 \pm 0.12$ \\
    $\textrm{30-ieee}$ & $29.83 \pm 5.39$ & $\textbf{0.23} \boldsymbol{\pm} \textbf{0.05}$ & $19.45 \pm 6.46$ & $2.39 \pm 0.12$ & $\textbf{0.06} \boldsymbol{\pm} \textbf{0.01}$ & $2.84 \pm 0.13$ \\
    $\textrm{39-epri}$ & $14.46 \pm 2.84$ & $\textbf{3.27} \boldsymbol{\pm} \textbf{0.18}$ & $15.09 \pm 2.92$ & $2.81 \pm 0.14$ & $\textbf{2.11} \boldsymbol{\pm} \textbf{0.07}$ & $3.24 \pm 0.19$ \\
    $\textrm{57-ieee}$ & $8.53 \pm 3.65$ & $\textbf{2.29} \boldsymbol{\pm} \textbf{0.15}$ & $9.80 \pm 4.50$ & $2.14 \pm 0.15$ & $\textbf{1.09} \boldsymbol{\pm} \textbf{0.17}$ & $2.35 \pm 0.22$ \\
    $\textrm{73-ieee-rts}$ & $36.85 \pm 1.53$ & $\textbf{31.69} \boldsymbol{\pm} \textbf{0.11}$ & $53.01 \pm 1.03$ & $1.31 \pm 0.14$ & $\textbf{0.35} \boldsymbol{\pm} \textbf{0.04}$ & $1.67 \boldsymbol{\pm} 0.13$ \\
    $\textrm{118-ieee}$ & $31.57 \pm 3.29$ & $\textbf{6.47} \boldsymbol{\pm} \textbf{0.20}$ & $39.85 \pm 7.85$ & $3.91 \pm 0.09$ & $\textbf{1.41} \boldsymbol{\pm} \textbf{0.09}$ & $4.34 \pm 0.27$ \\
    $\textrm{162-ieee-dtc}$ & $11.71 \pm 0.61$ & $\textbf{6.27} \boldsymbol{\pm} \textbf{0.18}$ & $11.81 \pm 0.60$ & $6.40 \pm 0.12$ & $\textbf{3.47} \boldsymbol{\pm} \textbf{0.11}$ & $5.55 \pm 0.14$ \\
    $\textrm{300-ieee}$ & $16.79 \pm 2.59$ & $\textbf{9.35} \boldsymbol{\pm} \textbf{0.15}$ & $46.63 \pm 8.50$ & $3.48 \pm 0.08$ & $\textbf{2.83} \boldsymbol{\pm} \textbf{0.08}$ & $5.01 \pm 1.34$ \\
    $\textrm{588-sdet}$ & $19.98 \pm 2.27$ & $\textbf{16.30} \boldsymbol{\pm} \textbf{0.24}$ & $22.48 \pm 0.95$ & $5.64 \pm 0.18$ & $\textbf{4.20} \boldsymbol{\pm} \textbf{0.07}$ & $15.51 \pm 2.25$ \\
    \bottomrule
    \end{tabular}
    %}
\end{table*}

One could argue that the improvement is due to the increased number of parameters, which did indeed approximately double (Table~\ref{tab:params_reg_local}).
However, comparing the performance of the two sets of global models, the difference seems to be marginal, highlighting again the utility of the fully connected component and confirming our suspicion of a lack of locality within this problem.

\begin{table*}[!ht]
\small
\caption{Number of parameters for local and extended global regression GNN models (fixed and varying topology)}
\label{tab:params_reg_local}
\def\na{---}
\centering
    %\resizebox{\columnwidth}{!}{
    \begin{tabular}{lrrr|rrrr}
    \toprule
    \multirow{2}{*}{Case} & \multicolumn{6}{c}{\# of parameters} \\ 
    \cmidrule(r){2-7}
    &$\mathcal{M}^{\textrm{GCN}}_{\textrm{local-3}}$ & $\mathcal{M}^{\textrm{CHC}}_{\textrm{local-3}}$ & $\mathcal{M}^{\textrm{GAT}}_{\textrm{local-3}}$ & $\mathcal{M}^{\textrm{GCN}}_{\textrm{global-4}}$ & $\mathcal{M}^{\textrm{CHC}}_{\textrm{global-4}}$ & $\mathcal{M}^{\textrm{GAT}}_{\textrm{global-4}}$ \\
    \midrule
    $\textrm{24-ieee-rts}$ & $2796$ & $3165$ & $2996$ & $6888$ & $7257$ & $7088$ \\
    $\textrm{30-ieee}$ & $796$ & $1045$ & $900$ & $1528$ & $1777$ & $1632$ \\
    $\textrm{39-epri}$ & $1355$ & $1629$ & $1493$ & $2935$ & $3209$ & $3073$ \\
    $\textrm{57-ieee}$ & $1541$ & $1935$ & $1703$ & $3151$ & $3545$ & $3313$ \\
    $\textrm{73-ieee-rts}$ & $20583$ & $21451$ & $21145$ & $57411$ & $58279$ & $57973$ \\
    $\textrm{118-ieee}$ & $27912$ & $28835$ & $28600$ & $53508$ & $54431$ & $54196$ \\
    $\textrm{162-ieee-dtc}$ & $9844$ & $10969$ & $10284$ & $17644$ & $18769$ & $18084$ \\
    $\textrm{300-ieee}$ & $87526$ & $89662$ & $88698$ & $170464$ & $172600$ & $171636$ \\
    $\textrm{588-sdet}$ & $220332$ & $224469$ & $222260$ & $432412$ & $436549$ & $434340$ \\
    \bottomrule
    \end{tabular}
    %}
\end{table*}

Finally, we also investigated the utility of using the nodal admittance matrix to express electrical distances within the power grid -- i.e. setting $k = 1$ in eq. $\eqref{edge-weights}$ --, rather than the simple binary adjacency matrix ($k = 0$). For this inherently more sophisticated approach, the results were in fact fairly consistent to those with $k = 0$ (a table summarising the MSE statistics for such models can be found in the \textit{Supplementary Materials}).
This is again in accordance with our suspicion that locality between input and output variables for this set of problems is rather limited, hence even more sophisticated measures of distance still cannot improve the performance of the GNNs.

\subsubsection{Classification}
In principle, the binding status of constraints could be predicted as nodal and edge features within a GNN framework.
However, based on our findings for the regression experiments (i.e. that the global strategy significantly outperforms the local one), we treated constraints only as global variables.
Classification performance is reported in terms of statistics of BCE of the test set, again based on 10 independent runs (Table~\ref{tab:error_clf_global}). Additional tables concerning the number of parameters as well as the training time for each model can be found in the \textit{Supplementary Materials}.

Here, the single-layer FCNN was observed to be even more dominant relative to the regression case.
Interestingly, for larger grids, it even outperforms the three-layer FCNN, which could be suffering from over-fitting as a consequence of increased flexibility. In general, we reach a similar conclusion as in the global regression setting, whereby the performance enhancements of the GNN classifiers are marginal respective to their practicality and computational limitations.
CHC and SC models perform similarly, but CHC remains the cheaper option with respect to the training time. Note that GAT was excluded from these experiments since it had already shown weak performance for the regression case relative to the other GNN models.

Although for brevity we only present the test set loss, we also note that we observed a greater precision than recall in virtually every instance.
This implies that the BCE objective is more sensitive to false positives.
In combination with the iterative feasibility test, which is more sensitive to false negative predictions, this can result in a significant increase in the computational cost of obtaining solutions~\cite{Robson2020}.
In order to fix this misalignment, one could either use a weighted BCE (with appropriate weights for the corresponding terms) or a meta-loss objective function~\cite{Jamei2019}~\cite{Robson2020}.

\begin{table*}[!ht]
\small
\caption{BCE statistics (mean and two-sided 95\% confidence intervals) of the test sets for global classification models (fixed topology)}
\label{tab:error_clf_global}
\def\na{---}
\centering
    %\resizebox{\textwidth}{!}{
    \begin{tabular}{lr|rrrrrr}
    \toprule
    \multirow{2}{*}{Case} & \multicolumn{7}{c}{BCE ($\times 10^{-2}$)} \\
    \cmidrule(r){2-8}
    & $\mathcal{M}^{\textrm{FCNN}}_{\textrm{global-3}}$ & $\mathcal{M}^{\textrm{FCNN}}_{\textrm{global-1}}$ & $\mathcal{M}^{\textrm{CNN}}_{\textrm{global-4}}$ & $\mathcal{M}^{\textrm{GCN}}_{\textrm{global-3}}$ & $\mathcal{M}^{\textrm{CHC}}_{\textrm{global-3}}$ & $\mathcal{M}^{\textrm{SC}}_{\textrm{global-3}}$ & $\mathcal{M}^{\textrm{GC}}_{\textrm{global-3}}$ \\
    \midrule
    $\textrm{24-ieee-rts}$ & $1.89 \pm 0.10$ & $3.58 \pm 0.12$ & $4.66 \pm 0.52$ & $6.93 \pm 0.65$ & $\textbf{3.14} \boldsymbol{\pm} \textbf{0.18}$ & $3.52 \pm 0.21$ & $3.42 \pm 0.33$ \\
    $\textrm{30-ieee}$ & $1.71 \pm 0.31$ & $5.14 \pm 0.65$ & $4.00 \pm 0.55$ & $8.76 \pm 1.24$ & $\textbf{3.58} \boldsymbol{\pm} \textbf{0.28}$ & $5.33 \pm 1.21$ & $4.98 \pm 0.73$ \\
    $\textrm{39-epri}$ & $3.61 \pm 0.12$ & $7.55 \pm 0.21$ & $13.84 \pm 0.22$ & $10.48 \pm 0.31$ & $\textbf{7.07} \boldsymbol{\pm} \textbf{0.15}$ & $8.07 \pm 0.26$ & $7.60 \pm 0.35$ \\
    $\textrm{57-ieee}$ & $1.67 \pm 0.14$ & $2.51 \pm 0.24$ & $2.51 \pm 0.29$ & $2.81 \pm 0.17$ & $2.34 \pm 0.18$ & $2.24 \pm 0.24$ & $\textbf{2.12} \boldsymbol{\pm} \textbf{0.18}$ \\
    $\textrm{73-ieee-rts}$ & $3.06 \pm 0.14$ & $4.34 \pm 0.10$ & $4.71 \pm 0.25$ & $6.28 \pm 0.24$ & $\textbf{3.34} \boldsymbol{\pm} \textbf{0.11}$ & $4.26 \pm 0.59$ & $4.08 \pm 0.89$ \\
    $\textrm{118-ieee}$ & $4.51 \pm 0.25$ & $6.19 \pm 0.21$ & $8.29 \pm 0.39$ & $7.86 \pm 0.32$ & $4.65 \pm 0.19$ & $\textbf{4.35} \boldsymbol{\pm} \textbf{0.21}$ & $4.40 \pm 0.20$ \\
    $\textrm{162-ieee-dtc}$ & $5.42 \pm 0.29$ & $6.27 \pm 0.15$ & $6.31 \pm 0.34$ & $8.32 \pm 0.19$ & $6.19 \pm 0.18$ & $\textbf{5.99} \boldsymbol{\pm} \textbf{0.17}$ & $6.18 \pm 0.18$ \\
    $\textrm{300-ieee}$ & $9.32 \pm 0.23$ & $\textbf{8.43} \boldsymbol{\pm} \textbf{0.14}$ & $10.97 \pm 0.29$ & $10.20 \pm 0.33$ & $8.86 \pm 0.19$ & $8.70 \pm 0.16$ & $8.65 \pm 0.21$ \\
    $\textrm{588-sdet}$ & $10.92 \pm 0.22$ & $\textbf{8.75} \boldsymbol{\pm} \textbf{0.14}$ & $12.13 \pm 0.45$ & $12.14 \pm 0.37$ & $11.38 \pm 0.21$ & $11.46 \pm 0.18$ & $10.92 \pm 0.14$ \\
    $\textrm{1354-pegase}$ & $11.99 \pm 0.18$ & $\textbf{10.56} \boldsymbol{\pm} \textbf{0.10}$ & $21.56 \pm 0.98$ & $17.14 \pm 0.44$ & $18.80 \pm 0.32$ & $18.43 \pm 0.93$ & $17.86 \pm 0.60$ \\
    $\textrm{2853-sdet}$ & $17.30 \pm 0.36$ & $\textbf{11.55} \boldsymbol{\pm} \textbf{0.04}$ & $37.88 \pm 1.59$ & $28.58 \pm 0.88$ & $31.83 \pm 0.33$ & $30.37 \pm 0.53$ & $33.47 \pm 0.61$ \\
    \bottomrule
    \end{tabular}
    %}
\end{table*}

\subsection{Varying topology}
We now focus our analysis toward the predictive performance of NN models trained (and tested) using data derived from power grids of size 24 -- 588 with varying topology. 
In these experiments, we modeled the \nmo{} line contingency and samples for a given grid differed not only in their input grid parameters but also in their topology.
For FCNN and CNN models, we used only grid parameters as inputs to predict the corresponding quantities of regression and classification, similarly to the fixed topology. We note that in theory, the input vector could be extended to include topological information, but it is rather cumbersome due to the quadratic scaling of the weighted adjacency matrix with system size.
For GNN models, however, the change in the topology can be naturally taken into account by passing the graph information of the sample along with the grid parameters. For the regression and classification approaches we have: $\mathrm{NN}_{\theta}^{\mathrm{reg}}(x_{i}, \mathbb{G}_{i}) = \hat{y}_{i}^{*}$ and $\mathrm{NN}_{\theta}^{\mathrm{clf}}(x_{i}, \mathbb{G}_{i}) = \hat{\mathcal{A}}_{i}$, where $x_{i}$ and $\mathbb{G}_{i}$ are the grid parameter vector and topology of the $i$-th sample, respectively.

\subsubsection{Regression}
We begin our discussion again by evaluating the global regression models (Table~\ref{tab:error_reg_global_contingency}). As expected, due to the larger effective parameter space, the regression performance using samples with varying topology decreases when compared to those with fixed topology for all cases and architectures (c.f. Table~\ref{tab:error_reg_global}).
A significant difference is that the best GNN models -- CHC in most cases -- outperforms both the single-layer and even the three-layer FCNN models (and CNN models too). This is resultant of the fact that in these models, any change in the network topology is ignored, whilst in the GNN architectures it is considered explicitly. This is a promising finding for applications of GNN models for predicting solutions of more sophisticated OPF problems including contingencies.

Interestingly, further investigations revealed that locality properties still play a marginal role in the predictive performance of GNNs: as for the fixed topology cases, local GNN models have a significantly weaker performance, which is subsequently restored by attaching a readout layer (Table~\ref{tab:error_reg_local_contingency}).

\begin{table*}[!ht]
\small
\caption{MSE statistics (mean and two-sided 95\% confidence intervals) of the test sets for global regression models with varying topology}
\label{tab:error_reg_global_contingency}
\def\na{---}
\centering
    \resizebox{\textwidth}{!}{
    \begin{tabular}{lr|rrrrrrr}
    \toprule
    \multirow{2}{*}{Case} & \multicolumn{8}{c}{MSE ($\times 10^{-3}$)} \\
    \cmidrule(r){2-9}
    & 
    $\mathcal{M}^{\textrm{FCNN}}_{\textrm{global-3}}$ & $\mathcal{M}^{\textrm{FCNN}}_{\textrm{global-1}}$ & $\mathcal{M}^{\textrm{CNN}}_{\textrm{global-4}}$ & $\mathcal{M}^{\textrm{GCN}}_{\textrm{global-3}}$ & $\mathcal{M}^{\textrm{CHC}}_{\textrm{global-3}}$ & $\mathcal{M}^{\textrm{SC}}_{\textrm{global-3}}$ & $\mathcal{M}^{\textrm{GC}}_{\textrm{global-3}}$ & $\mathcal{M}^{\textrm{GAT}}_{\textrm{global-3}}$ \\
    \midrule
    $\textrm{24-ieee-rts}$ & $1.27 \pm 0.18$ & $1.62 \pm 0.16$ & $1.42 \pm 0.17$ & $1.91 \pm 0.17$ & $\textbf{0.99} \boldsymbol{\pm} \textbf{0.08}$ & $1.42 \pm 0.18$ & $1.25 \pm 0.12$ & $1.65 \pm 0.13$ \\
    $\textrm{30-ieee}$ & $8.77 \pm 0.22$ & $8.39 \pm 0.19$ & $8.53 \pm 0.18$ & $8.68 \pm 0.16$ & $\textbf{0.23} \boldsymbol{\pm} \textbf{0.04}$ & $1.92 \pm 0.62$ & $0.66 \pm 0.08$ & $3.43 \pm 0.37$ \\
    $\textrm{39-epri}$ & $12.72 \pm 0.28$ & $12.09 \pm 0.22$ & $13.33 \pm 0.21$ & $12.56 \pm 0.24$ & $\textbf{3.31} \boldsymbol{\pm} \textbf{0.16}$ & $5.65 \pm 0.85$ & $4.23 \pm 0.23$ & $7.86 \pm 0.33$ \\
    $\textrm{57-ieee}$ & $4.34 \pm 0.12$ & $3.88 \pm 0.13$ & $4.01 \pm 0.12$ & $3.96 \pm 0.13$ & $\textbf{0.82} \boldsymbol{\pm} \textbf{0.08}$ & $2.81 \pm 0.72$ & $1.27 \pm 0.13$ & $2.43 \pm 0.16$ \\
    $\textrm{73-ieee-rts}$ & $0.85 \pm 0.05$ & $0.95 \pm 0.05$ & $1.01 \pm 0.04$ & $1.16 \pm 0.06$ & $\textbf{0.66} \boldsymbol{\pm} \textbf{0.07}$ & $0.92 \pm 0.06$ & $0.86 \pm 0.07$ & $1.24 \pm 0.18$ \\
    $\textrm{118-ieee}$ & $3.06 \pm 0.14$ & $2.59 \pm 0.12$ & $2.88 \pm 0.11$ & $2.86 \pm 0.12$ & $\textbf{1.15} \boldsymbol{\pm} \textbf{0.0}5$ & $1.78 \pm 0.12$ & $1.38 \pm 0.08$ & $2.66 \pm 0.34$ \\
    $\textrm{162-ieee-dtc}$ & $5.37 \pm 0.18$ & $4.38 \pm 0.17$ & $4.59 \pm 0.13$ & $5.81 \pm 0.15$ & $4.27 \pm 0.13$ & $5.29 \pm 0.14$ & $\textbf{3.95} \boldsymbol{\pm} \textbf{0.20}$ & $5.29 \pm 0.16$ \\
    $\textrm{300-ieee}$ & $3.24 \pm 0.08$ & $3.16 \pm 0.08$ & $4.02 \pm 0.27$ & $3.62 \pm 0.09$ & $\textbf{2.42} \boldsymbol{\pm} \textbf{0.07}$ & $2.79 \pm 0.07$ & $2.64 \pm 0.06$ & $3.72 \pm 0.08$ \\
    $\textrm{588-sdet}$ & $4.95 \pm 0.12$ & $4.02 \pm 0.12$ & $4.98 \pm 0.14$ & $4.63 \pm 0.14$ & $3.83 \pm 0.31$ & $4.16 \pm 0.13$ & $\textbf{3.36} \boldsymbol{\pm} \textbf{0.56}$ & $4.46 \pm 0.13$ \\
    \bottomrule
    \end{tabular}
   }
\end{table*}

\begin{table*}[!ht]
\small
\caption{MSE statistics (mean and two-sided 95\% confidence intervals) of the test sets for local and extended global regression GNN models (varying topology)}
\label{tab:error_reg_local_contingency}
\def\na{---}
\centering
    %\resizebox{\textwidth}{!}{
    \begin{tabular}{lrrr|rrr}
    \toprule
    \multirow{2}{*}{Case} & \multicolumn{6}{c}{MSE ($\times 10^{-3}$)} \\
    \cmidrule(r){2-7}
    & $\mathcal{M}^{\textrm{GCN}}_{\textrm{local-3}}$ & $\mathcal{M}^{\textrm{CHC}}_{\textrm{local-3}}$ & $\mathcal{M}^{\textrm{GAT}}_{\textrm{local-3}}$ & $\mathcal{M}^{\textrm{GCN}}_{\textrm{global-4}}$ & $\mathcal{M}^{\textrm{CHC}}_{\textrm{global-4}}$ & $\mathcal{M}^{\textrm{GAT}}_{\textrm{global-4}}$ \\
    \midrule
    $\textrm{24-ieee-rts}$ & $76.18 \pm 8.12$ & $\textbf{26.59} \boldsymbol{\pm} \textbf{0.42}$ & $71.32 \pm 9.76$ & $1.88 \pm 0.16$ & $\textbf{0.74} \boldsymbol{\pm} \textbf{0.12}$ & $1.57 \pm 0.11$ \\
    $\textrm{30-ieee}$ & $13.35 \pm 2.06$ & $\textbf{3.41} \boldsymbol{\pm} \textbf{0.08}$ & $16.55 \pm 6.37$ & $8.68 \pm 0.18$ & $\textbf{0.15} \boldsymbol{\pm} \textbf{0.02}$ & $2.13 \pm 0.37$ \\
    $\textrm{39-epri}$ & $46.47 \pm 8.69$ & $\textbf{4.47} \boldsymbol{\pm} \textbf{0.22}$ & $24.97 \pm 7.03$ & $12.51 \pm 0.21$ & $\textbf{2.97} \boldsymbol{\pm} \textbf{0.13}$ & $6.75 \pm 1.06$ \\
    $\textrm{57-ieee}$ & $9.13 \pm 2.97$ & $\textbf{2.09} \boldsymbol{\pm} \textbf{0.23}$ & $24.13 \pm 9.47$ & $4.11 \pm 0.12$ & $\textbf{0.66} \boldsymbol{\pm} \textbf{0.08}$ & $1.74 \pm 0.21$ \\
    $\textrm{73-ieee-rts}$ & $70.21 \pm 1.84$ & $\textbf{65.43} \boldsymbol{\pm} \textbf{0.06}$ & $99.92 \pm 7.04$ & $1.23 \pm 0.17$ & $\textbf{0.42} \boldsymbol{\pm} \textbf{0.04}$ & $1.23 \pm 0.26$ \\
    $\textrm{118-ieee}$ & $22.55 \pm 6.68$ & $\textbf{4.88} \boldsymbol{\pm} \textbf{0.05}$ & $35.95 \pm 4.88$ & $3.02 \pm 0.12$ & $\textbf{1.55} \boldsymbol{\pm} \textbf{0.12}$ & $2.97 \pm 0.34$ \\
    $\textrm{162-ieee-dtc}$ & $24.74 \pm 7.34$ & $\textbf{6.48} \boldsymbol{\pm} \textbf{0.34}$ & $19.65 \pm 7.61$ & $6.38 \pm 0.12$ & $\textbf{4.77} \boldsymbol{\pm} \textbf{0.16}$ & $6.57 \pm 1.08$ \\
    $\textrm{300-ieee}$ & $16.86 \pm 2.55$ & $\textbf{6.84} \boldsymbol{\pm} \textbf{0.19}$ & $50.52 \pm 7.97$ & $3.68 \pm 0.09$ & $\textbf{3.02} \boldsymbol{\pm} \textbf{0.13}$ & $4.67 \pm 0.92$ \\
    $\textrm{588-sdet}$ & $11.87 \pm 5.38$ & $\textbf{7.18} \boldsymbol{\pm} \textbf{0.18}$ & $22.62 \pm 0.19$ & $4.89 \pm 0.14$ & $\textbf{4.36} \boldsymbol{\pm} \textbf{0.13}$ & $6.96 \pm 0.98$ \\
    \bottomrule
    \end{tabular}
    %}
\end{table*}

\subsubsection{Classification}
For the classification models, we considered again only the global case (Table~\ref{tab:error_clf_global_contingency}). We note that due to the higher number of non-trivial constraints, the size of the NN models with varying topology differs from those with fixed topology (details are shown in the \textit{Supplementary Materials}). Therefore, unlike in the case of regression, we cannot compare directly the BCE statistics of experiments with fixed and varying topology.  
Nevertheless, in general, we found a similar trend to the global regression, i.e. the best performing GNN model (again, most often CHC) consistently outperforms the single-layer FCNN, the CNN and even the three-layer FCNN models. This means that applying GNN models is preferable over a significantly larger FCNN architecture for both OPF related regression and classification based problems with varying topology.

\begin{table*}[!ht]
\small
\caption{BCE statistics (mean and two-sided 95\% confidence intervals) of the test sets for global classification models with varying topology}
\label{tab:error_clf_global_contingency}
\def\na{---}
\centering
    %\resizebox{\textwidth}{!}{
    \begin{tabular}{lr|rrrrrr}
    \toprule
    \multirow{2}{*}{Case} & \multicolumn{7}{c}{BCE ($\times 10^{-2}$)} \\
    \cmidrule(r){2-8}
    & $\mathcal{M}^{\textrm{FCNN}}_{\textrm{global-3}}$ & $\mathcal{M}^{\textrm{FCNN}}_{\textrm{global-1}}$ & $\mathcal{M}^{\textrm{CNN}}_{\textrm{global-4}}$ & $\mathcal{M}^{\textrm{GCN}}_{\textrm{global-3}}$ & $\mathcal{M}^{\textrm{CHC}}_{\textrm{global-3}}$ & $\mathcal{M}^{\textrm{SC}}_{\textrm{global-3}}$ & $\mathcal{M}^{\textrm{GC}}_{\textrm{global-3}}$ \\
    \midrule
    $\textrm{24-ieee-rts}$ & $3.56 \pm 0.36$ & $3.19 \pm 0.17$ & $3.32 \pm 0.18$ & $3.43 \pm 0.16$ & $\textbf{1.44} \boldsymbol{\pm} \textbf{0.11}$ & $1.81 \pm 0.16$ & $1.67 \pm 0.16$ \\
    $\textrm{30-ieee}$ & $6.21 \pm 0.35$ & $6.19 \pm 0.32$ & $6.22 \pm 0.33$ & $6.59 \pm 0.37$ & $\textbf{3.03} \boldsymbol{\pm} \textbf{0.18}$ & $4.81 \pm 1.35$ & $4.43 \pm 0.17$ \\
    $\textrm{39-epri}$ & $8.66 \pm 0.19$ & $8.51 \pm 0.17$ & $9.06 \pm 0.21$ & $8.78 \pm 0.21$ & $\textbf{3.74} \boldsymbol{\pm} \textbf{0.12}$ & $5.71 \pm 0.86$ & $4.36 \pm 0.19$ \\
    $\textrm{57-ieee}$ & $5.34 \pm 0.24$ & $4.56 \pm 0.17$ & $4.65 \pm 0.18$ & $4.59 \pm 0.15$ & $\textbf{1.88} \boldsymbol{\pm} \textbf{0.07}$ & $3.48 \pm 0.93$ & $2.17 \pm 0.09$ \\
    $\textrm{73-ieee-rts}$ & $3.98 \pm 0.22$ & $3.87 \pm 0.16$ & $3.69 \pm 0.15$ & $4.18 \pm 0.28$ & $\textbf{2.25} \boldsymbol{\pm} \textbf{0.12}$ & $2.84 \pm 0.22$ & $2.92 \pm 0.21$ \\
    $\textrm{118-ieee}$ & $4.75 \pm 0.15$ & $3.95 \pm 0.11$ & $4.27 \pm 0.14$ & $4.28 \pm 0.12$ & $2.82 \pm 0.06$ & $3.42 \pm 0.14$ & $\textbf{2.79} \boldsymbol{\pm} \textbf{0.12}$ \\
    $\textrm{162-ieee-dtc}$ & $3.19 \pm 0.14$ & $2.66 \pm 0.06$ & $2.71 \pm 0.06$ & $3.23 \pm 0.06$ & $\textbf{2.17} \boldsymbol{\pm} \textbf{0.07}$ & $2.65 \pm 0.15$ & $2.66 \pm 0.07$ \\
    $\textrm{300-ieee}$ & $8.07 \pm 0.17$ & $7.35 \pm 0.11$ & $7.88 \pm 0.14$ & $7.43 \pm 0.17$ & $\textbf{6.38} \boldsymbol{\pm} \textbf{0.14}$ & $6.79 \pm 0.14$ & $6.74 \pm 0.17$ \\
    $\textrm{588-sdet}$ & $6.84 \pm 0.77$ & $5.91 \pm 0.07$ & $6.16 \pm 0.13$ & $5.87 \pm 0.12$ & $\textbf{5.12} \boldsymbol{\pm} \textbf{0.08}$ & $6.15 \pm 0.11$ & $5.91 \pm 0.08$ \\
    \bottomrule
    \end{tabular}
    %}
\end{table*}

\subsection{Locality properties}
Experimental results for the NN models indicated that the general assumption of locality may not be appropriate for this problem, i.e. there is only a weak -- or no existence of -- locality between load inputs and generator set-point outputs. To explore this relationship further, we carried out a sensitivity analysis that directly measures locality: for each synthetic grid, we iteratively perturbed each active load of 100 configurations by 1\% and recorded the absolute value of the relative change in voltage magnitude and active power injection of each generator (i.e. $\left| \frac{dV_{m}^{j}}{dP_{l}^{i}} \right|$ and $\left| \frac{dP_{g}^{j}}{dP_{l}^{i}} \right|$, where $P_{l}^{i}$ are the active loads with $i = 1, \dots, |\Load|$; and $V_{m}^{j}$ and $P_{g}^{j}$ are the voltage magnitude and injected active power of generators with $j = 1, \dots, |\Generator|$), as a function of neighbourhood order (i.e. the topological distance from the perturbed load). If a grid were to exhibit locality properties, one would expect a distinct negative correlation between the average of these quantities and the respective distance from the perturbed load within the graph domain.

\begin{figure*}
    \centerline{\includegraphics[width=\textwidth]{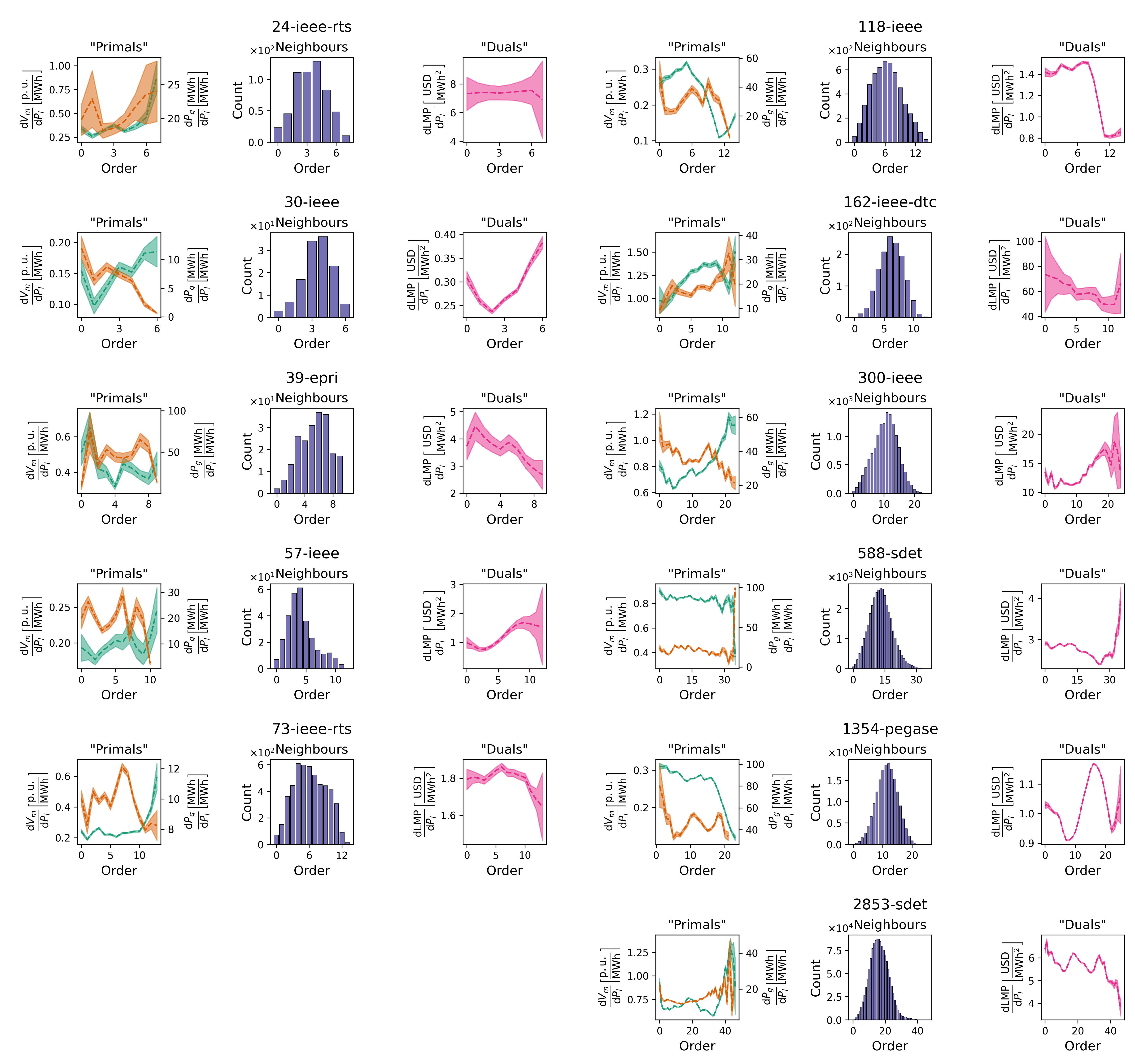}}
    \caption{Analysis of locality properties for each synthetic grid. Left and right panels show the average absolute value of the relative change (with two-sided 95\% confidence intervals) in voltage magnitude (green), injected active power (orange) and locational marginal prices (purple), respectively, as a function of the topological distance from the perturbed load. Center panels show the histogram of generators with respect to the neighbourhood order from loads.}
    \label{fig:locality}
\end{figure*}

The results of the sensitivity analysis are shown in the left panels of Figure~\ref{fig:locality}.
Although there are certain cases where either the voltage magnitude or active power injection show a weak anti-correlation with the topological distance, in general we found little evidence that the topology of the network influences the correlation between input and output variables. Plotting the distribution of generators as a function of distance from the perturbed load (middle panels of Figure~\ref{fig:locality}) suggests that this result should be of no surprise: as the system size increases, so does the average distance between the perturbed load and the generators in the system, which decreases the likelihood that nearby generators will balance corresponding demand (for apparent physical reasons such as generator capacity, line congestion etc.).

Finally, we also explored the existence of possible locality between grid inputs and the LMPs, which are functions of the duals (shadow prices) \cite{Singhal2019}. If a stronger locality property were to exist here this would be promising for using GNN models to predict electricity prices even with fixed topology \cite{Liu2021}. However, as shown in the right panels of Figure~\ref{fig:locality}, we found no evidence of locality for the LMP values either. 

\section{Conclusion}
With the potential to shift the entire computational effort to offline training, machine learning assisted OPF has become an increasingly interesting research direction.
Neural network based approaches are particularly promising as they can effectively model complex non-linear relationships between grid parameters and primal or dual variables of the underlying OPF problem.

Although most related works have applied fully connected neural networks so far, these networks scale relatively poorly with system size.
Therefore, incorporating topological information of the electricity grid into the inductive bias of some graph neural network is a sensible step towards reducing the number of NN parameters.

In this paper, we first provided a general framework of the most widely used end-to-end and hybrid techniques and showed that they can be considered as estimators of the OPF operator or function. In this sense, our framework could be readily extended to more sophisticated OPF problems, such as consideration of unit commitment or security constraints, as well as direct prediction of derived market signals (e.g. LMPs).

We then presented a systematic comparison of several NN architectures including FCNN, CNN and GNN models.
We found that for systems with fixed topology, an FCNN model has a comparable or even better predictive performance than global CNN and GNN models with similar number of parameters.
The moderate performance of the CNN model can be explained by the fact that it carries out convolutions in Euclidean space (instead of the graph domain).
We also identified that in the case of global GNN models, the readout layer plays a key role: constructing local models by removing their readout layer led to a significant decline in the predictive performance.

The results with fixed topology indicated that the required assumption of locality between grid parameters (inputs) and generator set-points (outputs) might not hold.
To validate the findings of the NN experiments, by carrying out a sensitivity analysis we showed that locality properties are indeed scarce between grid parameters and primal variables of the OPF.
Additionally, we found a similar lack of locality between grid parameters and LMPs.

Finally, we also performed a systematic comparison of NN models using varying topology of the samples.
In these experiments, we modeled the \nmo{} contingency of transmission lines in both the training and test sets.
We found that for such cases, global GNN architectures outperform FCNN and CNN models for both regression and classification based problems.
The reason is that although locality properties still play a limited role, GNN models could take the changes of the topology into account, which were completely neglected amongst FCNN and CNN models in our setup.
Although it might be possible to extend FCNN and CNN models' input by topology related features, it is definitely less straightforward than for GNN models, where this information is accounted for naturally.
This property of the GNN architectures therefore makes these models promising for realistic applications, especially for security constrained OPF problems.

\ifCLASSOPTIONcaptionsoff
  \newpage
\fi

\bibliographystyle{UTphys}

\bibliography{refs.bib}

\newpage
\includepdf[pages=-]{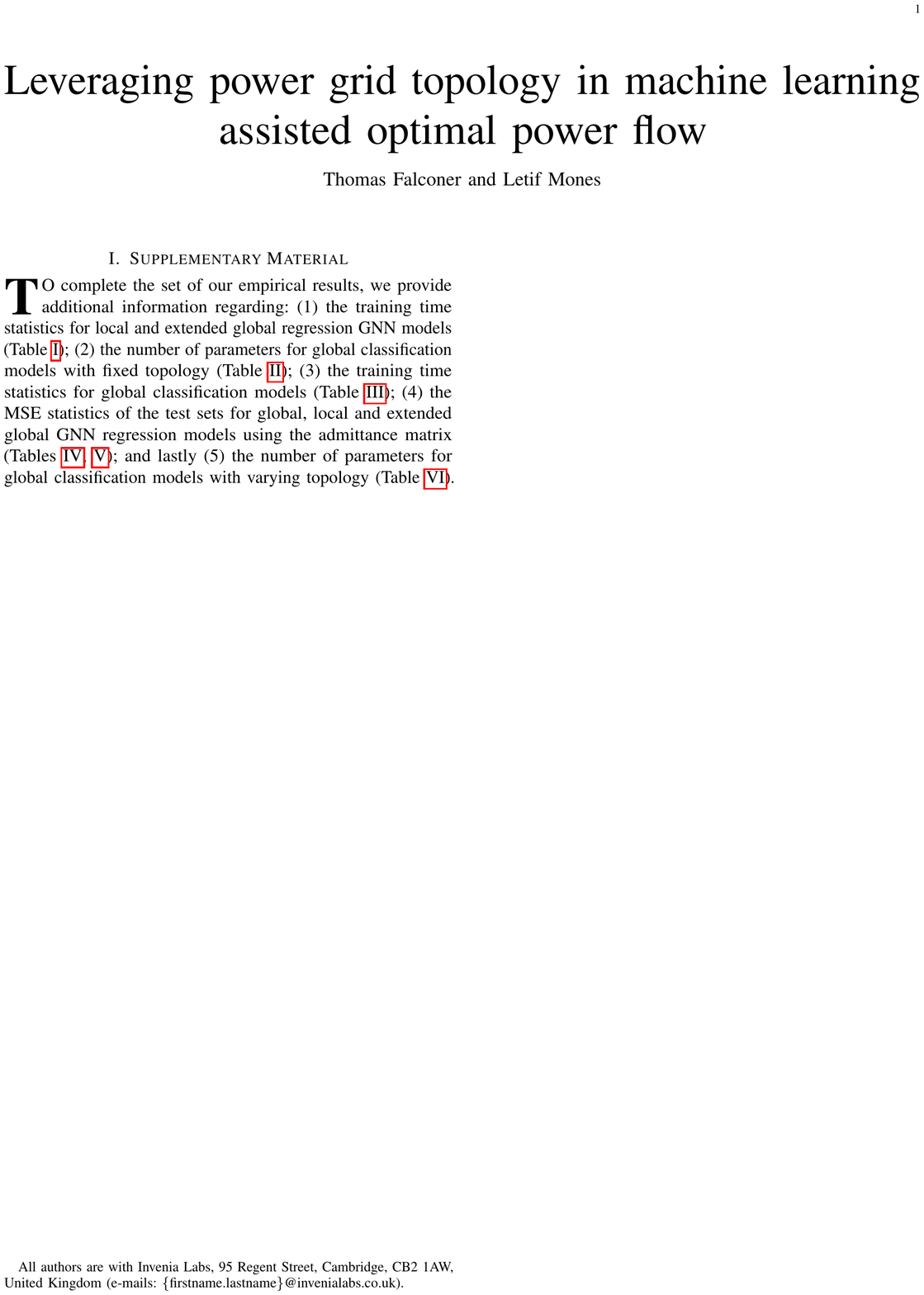}

\end{document}

% set url style
\urlstyle{tt}

\title{Leveraging power grid topology in machine learning assisted optimal power flow}
% note positions of commas and nonbreaking spaces $ ~ $ LaTeX will not break

\author{Thomas~Falconer
        and~Letif~Mones% <-this % stops a space
\thanks{All authors are with Invenia Labs, 95 Regent Street, Cambridge, CB2 1AW, United Kingdom (e-mails: \{firstname.lastname\}@invenialabs.co.uk).}% <-this % stops a space
}

\maketitle

\IEEEpeerreviewmaketitle

\section{Supplementary Material}

\IEEEPARstart{T}{o} complete the set of our empirical results, we provide additional information regarding: (1) the training time statistics for local and extended global regression GNN models (Table~\ref{tab:train_times_reg_local}); (2) the number of parameters for global classification models with fixed topology (Table~\ref{tab:params_clf_global}); (3) the training time statistics for global classification models (Table~\ref{tab:train_times_clf_global}); (4) the MSE statistics of the test sets for global, local and extended global GNN regression models using the admittance matrix (Tables~\ref{tab:error_reg_global_admittance}, \ref{tab:error_reg_local_admittance}); and lastly (5) the number of parameters for global classification models with varying topology (Table~\ref{tab:params_clf_global_contingency}).

\begin{table*}[!ht]
\small
\caption{Training time statistics (mean and two-sided 95\% confidence intervals) for local and extended global regression GNN models}
\label{tab:train_times_reg_local}
\def\na{---}
\centering
    %\resizebox{\textwidth}{!}{
    \begin{tabular}{lrrr|rrr}
    \toprule
    \multirow{2}{*}{Case} & \multicolumn{6}{c}{Training time ($\times 10^{2}$ s)} \\
    \cmidrule(r){2-7}
    & 
    $\mathcal{M}^{\textrm{GCN}}_{\textrm{local-3}}$ & $\mathcal{M}^{\textrm{CHC}}_{\textrm{local-3}}$ & $\mathcal{M}^{\textrm{GAT}}_{\textrm{local-3}}$ & $\mathcal{M}^{\textrm{GCN}}_{\textrm{global-4}}$ & $\mathcal{M}^{\textrm{CHC}}_{\textrm{global-4}}$ & $\mathcal{M}^{\textrm{GAT}}_{\textrm{global-4}}$ \\
    \midrule
    $\textrm{24-ieee-rts}$ & ${2.98} \pm {0.49}$ & $12.95 \pm 1.76$ & $6.30 \pm 1.38$ & ${8.25} \pm {1.55}$ & $14.92 \pm 2.25$ & $10.03 \pm 1.65$ \\
    $\textrm{30-ieee}$ & ${2.57} \pm {0.41}$ & $10.22 \pm 1.17$ & $6.54 \pm 1.31$ & ${9.57} \pm {2.36}$ & $15.86 \pm 1.72$ & $9.79 \pm 1.63$ \\
    $\textrm{39-epri}$ & ${2.60} \pm {0.54}$ & $11.03 \pm 0.91$ & $4.21 \pm 0.69$ & $13.18 \pm 2.14$ & $15.60 \pm 2.24$ & ${12.92} \pm {1.91}$ \\
    $\textrm{57-ieee}$ & ${2.47} \pm {0.77}$ & $9.47 \pm 0.71$ & $5.24 \pm 1.07$ & ${9.74} \pm {1.31}$ & $10.30 \pm 2.60$ & $10.87 \pm 1.92$ \\
    $\textrm{73-ieee-rts}$ & ${6.44} \pm {0.88}$ & $17.77 \pm 3.08$ & $12.12 \pm 3.67$ & ${11.58} \pm {1.69}$ & $23.40 \pm 3.17$ & $19.56 \pm 5.16$ \\
    $\textrm{118-ieee}$ & ${4.72} \pm {0.72}$ & $12.75 \pm 1.60$ & $9.44 \pm 1.65$ & $9.85 \pm 1.07$ & ${6.79} \pm {0.77}$ & $17.46 \pm 5.86$ \\
    $\textrm{162-ieee-dtc}$ & ${4.91} \pm {0.70}$ & $16.81 \pm 2.60$ & $10.05 \pm 1.52$ & $7.49 \pm 0.88$ & ${4.58} \pm {0.29}$ & $11.64 \pm 2.33$ \\
    $\textrm{300-ieee}$ & ${11.60} \pm {2.52}$ & $15.42 \pm 1.18$ & $20.00 \pm 4.91$ & $16.92 \pm 1.68$ & ${8.40} \pm {0.43}$ & $19.86 \pm 6.88$ \\
    $\textrm{588-sdet}$ & $43.33 \pm 9.31$ & ${27.33} \pm {2.18}$ & $75.15 \pm 9.95$ & $39.04 \pm 4.37$ & ${16.67} \pm {0.83}$ & $34.38 \pm 8.83$ \\
    \bottomrule
    \end{tabular}
    %}
\end{table*}

\begin{table*}[!ht]
\small
\caption{Number of parameters for global classification models (fixed topology)}
\label{tab:params_clf_global}
\def\na{---}
\centering
    %\resizebox{\columnwidth}{!}{
    \begin{tabular}{lr|rrrrrr}
    \toprule
    \multirow{2}{*}{Case} & \multicolumn{7}{c}{\# of parameters} \\ 
    \cmidrule(r){2-8}
    & $\mathcal{M}^{\textrm{FCNN}}_{\textrm{global-3}}$ & $\mathcal{M}^{\textrm{FCNN}}_{\textrm{global-1}}$ & $\mathcal{M}^{\textrm{CNN}}_{\textrm{global-4}}$ & $\mathcal{M}^{\textrm{GCN}}_{\textrm{global-3}}$ & $\mathcal{M}^{\textrm{CHC}}_{\textrm{global-3}}$ & $\mathcal{M}^{\textrm{SC}}_{\textrm{global-3}}$ & $\mathcal{M}^{\textrm{GC}}_{\textrm{global-3}}$ \\
    \midrule
    $\textrm{24-ieee-rts}$ & $3443$ & $833$ & $877$ & $1034$ & $1514$ & $1674$ & $1194$ \\
    $\textrm{30-ieee}$ & $3643$ & $244$ & $720$ & $359$ & $839$ & $999$ & $519$ \\
    $\textrm{39-epri}$ & $8120$ & $1659$ & $1617$ & $1075$ & $1555$ & $1715$ & $1235$ \\
    $\textrm{57-ieee}$ & $13301$ & $1150$ & $1398$ & $815$ & $1295$ & $1455$ & $975$ \\
    $\textrm{73-ieee-rts}$ & $30251$ & $7203$ & $6125$ & $7095$ & $7575$ & $7735$ & $7255$ \\
    $\textrm{118-ieee}$ & $63807$ & $9954$ & $9366$ & $10712$ & $12536$ & $13144$ & $11320$ \\
    $\textrm{162-ieee-dtc}$ & $128323$ & $24050$ & $21974$ & $24808$ & $26632$ & $27240$ & $25416$ \\
    $\textrm{300-ieee}$ & $494219$ & $114791$ & $107739$ & $115549$ & $117373$ & $117981$ & $116157$ \\
    $\textrm{588-sdet}$ & $1460783$ & $171842$ & $166590$ & $176688$ & $178512$ & $179120$ & $177296$ \\
    $\textrm{1354-pegase}$ & $7486643$ & $734139$ & $724971$ & $734897$ & $736721$ & $737329$ & $735505$ \\
    $\textrm{2853-sdet}$ & $43369442$ & $9690486$ & $9619758$ & $9759164$ & $9760988$ & $9761596$ & $9759772$ \\
    \bottomrule
    \end{tabular}
    %}
\end{table*}

\begin{table*}[!ht]
\small
\caption{Training time statistics (mean and two-sided 95\% confidence intervals) for global classification models}
\label{tab:train_times_clf_global}
\def\na{---}
\centering
    %\resizebox{\textwidth}{!}{
    \begin{tabular}{lr|rrrrrr}
    \toprule
    \multirow{2}{*}{Case} & \multicolumn{7}{c}{Training time ($\times 10^{2}$ s)} \\
    \cmidrule(r){2-8}
    & $\mathcal{M}^{\textrm{FCNN}}_{\textrm{global-3}}$ & $\mathcal{M}^{\textrm{FCNN}}_{\textrm{global-1}}$ & $\mathcal{M}^{\textrm{CNN}}_{\textrm{global-4}}$ & $\mathcal{M}^{\textrm{GCN}}_{\textrm{global-3}}$ & $\mathcal{M}^{\textrm{CHC}}_{\textrm{global-3}}$ & $\mathcal{M}^{\textrm{SC}}_{\textrm{global-3}}$ & $\mathcal{M}^{\textrm{GC}}_{\textrm{global-3}}$ \\
    \midrule
    $\textrm{24-ieee-rts}$ & $0.55 \pm 0.08$ & $4.21 \pm 0.02$ & ${0.79} \pm {0.05}$ & $9.14 \pm 2.29$ & $10.36 \pm 1.46$ & $7.87 \pm 0.62$ & $10.50 \pm 1.62$ \\
    $\textrm{30-ieee}$ & $0.36 \pm 0.04$ & $3.87 \pm 0.44$ & ${0.96} \pm {0.14}$ & $5.23 \pm 1.22$ & $11.35 \pm 1.49$ & $7.64 \pm 1.71$ & $8.32 \pm 1.35$ \\
    $\textrm{39-epri}$ & $0.51 \pm 0.06$ & $3.32 \pm 0.48$ & ${0.70} \pm {0.14}$ & $14.50 \pm 2.71$ & $11.81 \pm 1.37$ & $9.36 \pm 1.44$ & $10.94 \pm 1.68$ \\
    $\textrm{57-ieee}$ & $0.20 \pm 0.02$ & $1.98 \pm 0.73$ & ${0.49} \pm {0.04}$ & $9.29 \pm 1.79$ & $4.96 \pm 0.45$ & $6.37 \pm 1.10$ & $5.61 \pm 0.54$ \\
    $\textrm{73-ieee-rts}$ & $0.27 \pm 0.03$ & $4.44 \pm 0.16$ & ${1.06} \pm {0.30}$ & $12.92 \pm 2.53$ & $10.71 \pm 1.42$ & $9.45 \pm 1.52$ & $8.99 \pm 1.22$ \\
    $\textrm{118-ieee}$ & $0.20 \pm 0.01$ & $1.85 \pm 0.40$ & ${0.59} \pm {0.10}$ & $15.21 \pm 4.01$ & $4.62 \pm 0.28$ & $5.05 \pm 0.25$ & $4.36 \pm 0.39$ \\
    $\textrm{162-ieee-dtc}$ & $0.18 \pm 0.01$ & $0.81 \pm 0.09$ & ${0.52} \pm {0.05}$ & $7.86 \pm 0.87$ & $3.37 \pm 0.18$ & $4.37 \pm 0.41$ & $3.44 \pm 0.19$ \\
    $\textrm{300-ieee}$ & $0.15 \pm 0.00$ & $0.58 \pm 0.02$ & ${0.34} \pm {0.03}$ & $5.29 \pm 0.53$ & $3.44 \pm 0.11$ & $4.53 \pm 0.23$ & $2.81 \pm 0.15$ \\
    $\textrm{588-sdet}$ & $0.14 \pm 0.00$ & $0.50 \pm 0.02$ & ${0.23} \pm {0.01}$ & $4.00 \pm 0.41$ & $3.77 \pm 0.07$ & $5.05 \pm 0.19$ & $2.94 \pm 0.10$ \\
    $\textrm{1354-pegase}$ & $0.27 \pm 0.00$ & $0.50 \pm 0.02$ & ${0.20} \pm {0.00}$ & $2.61 \pm 0.13$ & $4.52 \pm 0.09$ & $7.07 \pm 0.17$ & $2.82 \pm 0.08$ \\
    $\textrm{2853-sdet}$ & $1.32 \pm 0.01$ & $0.65 \pm 0.02$ & ${0.40} \pm {0.01}$ & $8.53 \pm 0.19$ & $12.61 \pm 0.11$ & $17.26 \pm 0.44$ & $9.28 \pm 0.12$ \\
    \bottomrule
    \end{tabular}
    %}
\end{table*}

\begin{table*}[!ht]
\small
\caption{MSE statistics (mean and two-sided 95\% confidence intervals) of the test sets for global GNN regression models using admittance matrix (fixed topology)}
\label{tab:error_reg_global_admittance}
\def\na{---}
\centering
    %\resizebox{\textwidth}{!}{
    \begin{tabular}{lr|rrrrr}
    \toprule
    \multirow{2}{*}{Case} & \multicolumn{6}{c}{MSE ($\times 10^{-3}$)} \\
    \cmidrule(r){2-7}
    & 
    $\mathcal{M}^{\textrm{FCNN}}_{\textrm{global-3}}$ &  $\mathcal{M}^{\textrm{GCN}}_{\textrm{global-3}}$ & $\mathcal{M}^{\textrm{CHC}}_{\textrm{global-3}}$ & $\mathcal{M}^{\textrm{SC}}_{\textrm{global-3}}$ & $\mathcal{M}^{\textrm{GC}}_{\textrm{global-3}}$ & $\mathcal{M}^{\textrm{GAT}}_{\textrm{global-3}}$ \\
    \midrule
    $\textrm{24-ieee-rts}$ & $0.18 \pm 0.02$ & $2.48 \pm 0.12$ & $0.69 \pm 0.05$ & $0.89 \pm 0.07$ & $1.71 \pm 0.34$ & $2.78 \pm 0.18$ \\
    $\textrm{30-ieee}$ & $0.05 \pm 0.01$ & $1.41 \pm 0.28$ & $0.08 \pm 0.01$ & $0.11 \pm 0.03$ & $0.51 \pm 0.12$ & $2.95 \pm 0.22$ \\
    $\textrm{39-epri}$ & $0.89 \pm 0.10$ & $4.44 \pm 0.12$ & $2.42 \pm 0.05$ & $2.41 \pm 0.14$ & $4.86 \pm 0.95$ & $4.75 \pm 0.29$ \\
    $\textrm{57-ieee}$ & $0.52 \pm 0.11$ & $1.91 \pm 0.15$ & $1.25 \pm 0.13$ & $1.31 \pm 0.13$ & $1.67 \pm 0.15$ & $2.29 \pm 0.16$ \\
    $\textrm{73-ieee-rts}$ & $0.21 \pm 0.07$ & $1.37 \pm 0.26$ & $0.61 \pm 0.02$ & $0.67 \pm 0.02$ & $1.15 \pm 0.13$ & $1.83 \pm 0.11$ \\
    $\textrm{118-ieee}$ & $0.39 \pm 0.03$ & $2.12 \pm 0.07$ & $1.26 \pm 0.06$ & $1.27 \pm 0.07$ & $1.47 \pm 0.12$ & $2.43 \pm 0.12$ \\
    $\textrm{162-ieee-dtc}$ & $2.61 \pm 0.10$ & $3.75 \pm 0.11$ & $3.07 \pm 0.09$ & $2.87 \pm 0.13$ & $3.55 \pm 0.22$ & $4.84 \pm 0.22$ \\
    $\textrm{300-ieee}$ & $2.06 \pm 0.06$ & $3.96 \pm 0.16$ & $2.42 \pm 0.05$ & $2.35 \pm 0.07$ & $3.19 \pm 0.23$ & $3.59 \pm 0.24$ \\
    $\textrm{588-sdet}$ & $2.56 \pm 0.06$ & $3.66 \pm 0.07$ & $3.18 \pm 0.05$ & $3.21 \pm 0.08$ & $10.21 \pm 2.51$ & $5.01 \pm 0.24$ \\
    $\textrm{1354-pegase}$ & $0.83 \pm 0.12$ & $2.09 \pm 0.11$ & $1.43 \pm 0.09$ & $1.36 \pm 0.09$ & $2.64 \pm 0.11$ & $2.51 \pm 0.14$ \\
    $\textrm{2853-sdet}$ & $5.99 \pm 0.16$ & $11.54 \pm 0.48$ & $9.03 \pm 0.29$ & $8.35 \pm 0.13$ & $13.98 \pm 0.44$ & $11.15 \pm 0.59$ \\
    \bottomrule
    \end{tabular}
   %}
\end{table*}

\begin{table*}[!ht]
\small
\caption{MSE statistics (mean and two-sided 95\% confidence intervals) of the test sets for local and extended global regression GNN models using admittance matrix (fixed topology)}
\label{tab:error_reg_local_admittance}
\def\na{---}
\centering
    %\resizebox{\textwidth}{!}{
    \begin{tabular}{lrrr|rrr}
    \toprule
    \multirow{2}{*}{Case} & \multicolumn{6}{c}{MSE ($\times 10^{-3}$)} \\
    \cmidrule(r){2-7}
    & $\mathcal{M}^{\textrm{GCN}}_{\textrm{local-3}}$ & $\mathcal{M}^{\textrm{CHC}}_{\textrm{local-3}}$ & $\mathcal{M}^{\textrm{GAT}}_{\textrm{local-3}}$ & $\mathcal{M}^{\textrm{GCN}}_{\textrm{global-4}}$ & $\mathcal{M}^{\textrm{CHC}}_{\textrm{global-4}}$ & $\mathcal{M}^{\textrm{GAT}}_{\textrm{global-4}}$ \\
    \midrule
    $\textrm{24-ieee-rts}$ & $60.26 \pm 0.53$ & $25.89 \pm 0.14$ & $67.83 \pm 9.26$ & $2.16 \pm 0.08$ & $0.46 \pm 0.04$ & $2.56 \pm 0.15$ \\
    $\textrm{30-ieee}$ & $8.67 \pm 1.05$ & $0.52 \pm 0.14$ & $28.47 \pm 9.82$ & $1.17 \pm 0.38$ & $0.11 \pm 0.02$ & $2.85 \pm 0.22$ \\
    $\textrm{39-epri}$ &$16.12 \pm 6.49$ & $3.86 \pm 0.16$ & $13.94 \pm 2.79$ & $3.14 \pm 0.19$ & $2.21 \pm 0.08$ & $3.08 \pm 0.22$ \\
    $\textrm{57-ieee}$ & $4.85 \pm 0.21$ & $2.25 \pm 0.37$ & $8.87 \pm 3.25$ & $1.75 \pm 0.14$ & $1.26 \pm 0.21$ & $2.41 \pm 0.22$ \\
    $\textrm{73-ieee-rts}$ & $43.09 \pm 6.61$ & $31.58 \pm 0.04$ & $57.05 \pm 3.62$ & $0.88 \pm 0.03$ & $0.33 \pm 0.03$ & $2.32 \pm 1.33$ \\
    $\textrm{118-ieee}$ & $25.37 \pm 4.74$ & $7.75 \pm 0.19$ & $35.44 \pm 1.51$ & $2.65 \pm 0.11$ & $1.45 \pm 0.08$ & $4.66 \pm 0.26$\\
    $\textrm{162-ieee-dtc}$ & $11.19 \pm 0.18$ & $8.29 \pm 0.16$ & $11.51 \pm 0.13$ & $4.12 \pm 0.16$ & $3.33 \pm 0.11$ & $5.95 \pm 1.12$ \\
    $\textrm{300-ieee}$ & $11.56 \pm 0.46$ & $9.51 \pm 0.07$ & $78.56 \pm 9.67$ & $4.35 \pm 0.17$ & $2.71 \pm 0.06$ & $5.29 \pm 1.63$ \\
    $\textrm{588-sdet}$ & $21.11 \pm 1.15$ & $16.28 \pm 0.19$ & $28.56 \pm 8.91$ & $3.87 \pm 0.06$ & $3.83 \pm 0.12$ & $71.49 \pm 9.76$ \\
    \bottomrule
    \end{tabular}
    %}
\end{table*}

\begin{table*}[!ht]
\small
\caption{Number of parameters for global classification models (varying topology)}
\label{tab:params_clf_global_contingency}
\def\na{---}
\centering
    %\resizebox{\columnwidth}{!}{
    \begin{tabular}{lr|rrrrrr}
    \toprule
    \multirow{2}{*}{Case} & \multicolumn{7}{c}{\# of parameters} \\ 
    \cmidrule(r){2-8}
    & $\mathcal{M}^{\textrm{FCNN}}_{\textrm{global-3}}$ & $\mathcal{M}^{\textrm{FCNN}}_{\textrm{global-1}}$ & $\mathcal{M}^{\textrm{CNN}}_{\textrm{global-4}}$ & $\mathcal{M}^{\textrm{GCN}}_{\textrm{global-3}}$ & $\mathcal{M}^{\textrm{CHC}}_{\textrm{global-3}}$ & $\mathcal{M}^{\textrm{SC}}_{\textrm{global-3}}$ & $\mathcal{M}^{\textrm{GC}}_{\textrm{global-3}}$ \\
    \midrule
    $\textrm{24-ieee-rts}$ & $14372$ & $4263$ & $2067$ & $4324$ & $4804$ & $4964$ & $4484$ \\
    $\textrm{30-ieee}$ & $4740$ & $915$ & $1083$ & $700$ & $1180$ & $1340$ & $860$ \\
    $\textrm{39-epri}$ & $13661$ & $4345$ & $3283$ & $2435$ & $2915$ & $3075$ & $2595$ \\
    $\textrm{57-ieee}$ & $17233$ & $3680$ & $3180$ & $2091$ & $2571$ & $2731$ & $2251$ \\
    $\textrm{73-ieee-rts}$ & $77552$ & $25431$ & $20137$ & $24455$ & $24935$ & $25095$ & $24615$ \\
    $\textrm{118-ieee}$ & $112037$ & $34839$ & $31311$ & $35597$ & $37421$ & $38029$ & $36205$ \\
    $\textrm{162-ieee-dtc}$ & $171139$ & $47450$ & $42782$ & $48208$ & $50032$ & $50640$ & $48816$ \\
    $\textrm{300-ieee}$ & $600416$ & $169482$ & $158790$ & $170240$ & $172064$ & $172672$ & $170848$ \\
    $\textrm{588-sdet}$ & $1866920$ & $423720$ & $409908$ & $434558$ & $436382$ & $436990$ & $435166$ \\
    \bottomrule
    \end{tabular}
    %}
\end{table*}

\ifCLASSOPTIONcaptionsoff
  \newpage
\fi